\documentclass{ws-ijwmip}
\usepackage{graphicx,subfigure}
\usepackage{mathrsfs,amsmath,amsfonts,amssymb,bm}
\usepackage{multirow}
\usepackage{algorithm,algorithmic}
\usepackage{color,lineno}
\usepackage{bm}
\usepackage[super]{cite}

\begin{document}
\markboth{Y.Q. Xu et al.}{Federated Covariate Shift Adaptation for Missing Target Output Values}

\catchline{}{}{}{}{}

\title{Federated Covariate Shift Adaptation for Missing Target Output Values}

\author{Yaqian Xu}

\address{International Institute of Finance, School of Management,\\ University of Science and Technology of China,\\
Hefei, 230026,
P. R. China\\
xyq211@mail.ustc.edu.cn}

\author{Wenquan Cui}

\address{International Institute of Finance, School of Management,\\ University of Science and Technology of China,\\
Hefei, 230026,
P. R. China\\
wqcui@ustc.edu.cn}

\author{Jianjun Xu}

\address{International Institute of Finance, School of Management,\\ University of Science and Technology of China,\\
Hefei, 230026,
P. R. China\\
xjj1994@mail.ustc.edu.cn}

\author{Haoyang Cheng}

\address{College of Electrical and Information Engineering,\\ Quzhou University,\\
Quzhou, 324000,
P. R. China\\
chyling@mail.ustc.edu.cn}

\maketitle

\begin{history}
\received{(Day Month Year)}
\revised{(Day Month Year)}
\accepted{(Day Month Year)}
\published{(Day Month Year)}
\end{history}

\begin{abstract}
  The most recent multi-source covariate shift algorithm is an efficient hyperparameter optimization algorithm for missing target output. In this paper, we extend this algorithm to the framework of federated learning. For data islands in federated learning and covariate shift adaptation, we propose the federated domain adaptation estimate of the target risk which is asymptotically unbiased with a desirable asymptotic variance property. We construct a weighted model for the target task and propose the federated covariate shift adaptation algorithm which works preferably in our setting. The efficacy of our method is justified both theoretically and empirically.
\end{abstract}

\keywords{Federated learning; covariate shift; multi-source; hyperparameter optimization.}

\ccode{AMS Subject Classification: 62J07, 62-08}

\section{Introduction}\label{SecIntro}

Covariate shift adaptation has been a pivotal part of transfer learning.\cite{2007Covariate} It is a prevalent setting for machine learning in which the feature distributions differ between the train (source) and test (target) domains, while the conditional distributions of the output variable given the feature variable remain unchanged,~see Ref.~\refcite{2000Improving}. In this setting, when the output values of the target data are missing, the multi-source covariate shift (MS-CS)\cite{nomura2021efficient} algorithm can use multiple relevant source datasets whose output values are available to predict the target task.

One critical, but often overlooked assumption in MS-CS is that we can merge data from all sources as a training set. However, this assumption is not satisfied in many cases. For example, we need to predict the disease scores for patients in a new (target) hospital, but we only have historical medical outcome data from some source hospitals other than the target one. Furthermore, to protect patient privacy, each source hospital does not allow its data to leave its local area or to be disclosed, resulting in data islands. In this case, the MS-CS algorithm is infeasible. Fortunately, this issue can be well-solved in federated learning since it provides a privacy protection mechanism and allows us to learn and save locally at each node rather than share data or parameters,~see Ref.~\refcite{2018Aby,2017Practical,2017SecureML} and \refcite{2019Federated}.

In this paper, we extend the MS-CS algorithm to the federated learning framework. The difficulty in conducting MS-CS under federated learning is that the variance reduced estimate\cite{nomura2021efficient} of the target risk is inaccessible due to data islands between the sources. It is thus essential to construct a new estimate of the target risk in our setting.

To solve this problem, we first formulate the federated covariate shift (FedCS) setting. For covariate shift adaptation in the FedCS setting, we propose the federated importance weighting estimate (FedIWE) of the target risk which is asymptotically unbiased. Based on FedIWE, we further propose the federated domain adaptation estimate (FedDAE) of the target risk, which has the smallest asymptotic variance in a class of asymptotically unbiased estimates of the target risk. These estimates can well satisfy the requirement that every source does not allow its data to leave its local area or be disclosed. We provide theoretical analysis and proofs for all properties. The estimate of the hyperparameter is obtained by minimizing FedDAE, and then all source models are determined. We construct a weighted model by weighting all source models for the target task and present the error bound of the weighted model. In order to describe the steps of our proposed method in the specific implementation, we propose the federated covariate shift adaptation (FedCSA) algorithm. We execute the FedCSA algorithm on simulated data and real data, and the experimental results demonstrate that our proposed method is effective.

\textbf{Related Work}. Covariate shift adaptation aims to evaluate the performance of models for the target task using only a relevant single source dataset,~see Ref.~\refcite{2000Improving,2007Covariate,2010A,2019Towards} and \refcite{2010Cross}. Combining additional source information can improve the efficiency of searching hyperparameters and obtain a better solution with less computation,~see Ref.~\refcite{bonilla2007multi,feurer2018practical} and \refcite{NEURIPS2018_14c879f3}. Elvira \emph{et al}.\cite{2015Efficient} and Sugiyama \emph{et al}.\cite{2007Covariate} offer an estimate of ground-truth model performance based on importance sampling that is guaranteed unbiased in theory,~see Ref.~\refcite{2007Covariate} and \refcite{2018Conditional}. However, the variance is unbounded. Deep embedded validation combines the control variates method to reduce the variance,~see Ref.~\refcite{2019Towards} and \refcite{Lemieux2017}. Nomura and Saito\cite{nomura2021efficient} propose the variance reduced estimate in the case where the labels of the target data are completely missing under the MS-CS setting. Previous studies have often assumed only a single source or multi sources without data islands. This paper, on the other hand, focuses more on the real situation with the problem of data islands.

Another related field is federated learning,~see Ref.~\refcite{2018Aby,2017Practical,2017SecureML} and \refcite{2019Federated}. CryptoNet\cite{2016CryptoNets} improves the efficiency of data encryption and the performance of federated learning. Bonawitz \emph{et al}.\cite{2017Practical} proposes the aggregation scheme that updates the machine learning model under the federated learning framework. SecureML\cite{2017SecureML} supports the cooperative privacy protection training in the multi-client federated learning system. All of these methods learn a single global model. On the contrary, federated multi-task learning\cite{2017Federated} learns a separate model for each node. Liu \emph{et al.}\cite{2020secure} propose a semi-supervised federated transfer learning method under privacy protection. In their settings, the target output values are available. Federated adversarial domain adaptation proposes an efficient adaptation algorithm that can be applied to the federated setting based on adversarial adaptation and representation disentanglement,~see Ref.~\refcite{2019FederatedP}. These approaches are capable of resolving data island issues, and we are concentrating our efforts on extending MS-CS to the federated learning framework.

\textbf{Contributions}. Our contributions are summarized as follows:
\begin{itemize}
\item We formulate the FedCS setting which extends the MS-CS setting to the framework of federated learning.
\item We propose FedIWE which is asymptotically unbiased for the target risk in the FedCS setting.
\item We further propose FedDAE which has the smallest asymptotic variance in a class of asymptotically unbiased estimates of the target risk.
\item We construct a weighted model for the target task by weighting all source models and present the error bound of the weighted model.
\item We propose the FedCSA algorithm, which we have empirically proven to work well in the FedCS setting.
\end{itemize}

The remainder of this paper is organized as follows. In Sec.~\ref{SecMethod}, we formulate the FedCS setting and propose FedIWE for covariate shift. We futher propose FedDAE which has the smallest asymptotic variance in a class of asymptotically unbiased estimates of the target ris. We construct a weighted model for the target task and propose the FedCSA algorithm. The relevant theoretical proofs are provided in the appendix. Sec.~\ref{SecSimul} and Sec.~\ref{SecReal} justify the efficacy of our method through experiments on the simulation data and a real dataset, respectively. Sec.~\ref{SecCon} concludes.

\section{Method}\label{SecMethod}

In this paper, we call the data owner whose output values are available the source, and a new data owner whose output values are missing the target. We consider the case of $K$ sources and one target. Each source and target has a domain and a task,~see Ref.~\refcite{2010A}.

\subsection{FedCS setting}

Before formulating the FedCS setting, we first introduce some notations. Let $\mathcal{X}\subset\mathbb{R}^d$ be the d-dimensional continuous bounded feature space and $\mathcal{Y}\subset\mathbb{R}$ be the bounded real-valued output space. We use $P(\bm x)$ to denote the probability distribution function of the feature variable $\bm X\in\mathcal{X}$. Denote the $j$th source and target domains as $\mathcal{D}_{j}=\{\mathcal{X},P_j(\bm{\bm x})\}$ and $\mathcal{D}_T=\{\mathcal{X},P_T(\bm{x})\}$. The $j$th source and target tasks are denoted by $\mathcal{T}_j=\{\mathcal{Y},f_j(\cdot)\}$ and $\mathcal{T}_T=\{\mathcal{Y},f_T(\cdot)\}$, where $f(\cdot)$ is the function which predicts the output value using the feature instance. We use $p_j(\bm x)$, $p_T(\bm x)$, $p_j(\bm x,y)$ and $p_T(\bm x,y)$ to denote the corresponding feature probability density functions and joint probability density functions, respectively. More specifically, we denote the i.i.d. dataset of the $j$th source as $D_{j}=\{(\bm{x}_{j,i},y_{j,i})\}_{i=1}^{n_{j}}$, where $\bm{x}_{j,i}\in\mathcal{X}$ is the feature instance and $y_{j,i}\in\mathcal{Y}$ is the corresponding output value. Similarly, we denote the i.i.d. dataset of the target as $D_T=\{(\bm{x}_{T,i})\}_{i=1}^{n_T}$. In most cases, $0<n_T\ll\sum_{j=1}^K n_j$.

Next, we present the detail of the FedCS setting and two assumptions for covariate shift are described as follows:

\begin{assumption}\label{assu01}
The source domains have support for the target domain, while the marginal feature distributions are different, i.e., $p_T(\bm{x})>0\Rightarrow p_{j}(\bm{x})>0$, $p_T(\bm x)\neq p_j(\bm x)$, $\forall \bm{x}\in\mathcal{X}$, $\forall j=1,2,\cdots,K$.
\end{assumption}

\begin{assumption}\label{assu02}
Conditional output distributions remain the same between the target and all the sources, i.e., $p_T(y|\bm{x})=p_{j}(y|\bm{x}),j =1,2,\cdots,K$, where $p_T(y|\bm x)$ and $p_j(y|\bm x)$ denote the conditional output density functions of the target and $j$th source, respectively.
\end{assumption}

Assumption \ref{assu01} is a commonly adopted condition in covariate shift, and Assumption \ref{assu02} is required to ensure that the source data can be used to predict the target output values,~see Ref.~\refcite{2000Improving}. Privacy protection, in addition to covariate shift, needs to be considered in the FedCS setting. In our paper, every source does not allow its data to leave its local area or to be disclosed, and the output values of the target data are completely missing. The purpose of this paper is to learn a parametric model that can accurately predict the output values of the target data in the FedCS setting.

\subsection{Estimation of the target risk function}
To obtain an accurate parametric model, we need to find the optimal hyperparameter $\bm{\theta}^\star$ with respect to the target distribution:   \begin{equation}\label{01}
\bm{\theta}^{\star}=\underset{\bm{\theta} \in \Theta}{\arg \min } f_T(\bm\theta)=\underset{\bm{\theta} \in \Theta}{\arg \min } \mathbb{E}_{(\boldsymbol{X}, Y) \sim p_{T}(\boldsymbol{x}, y)}\Big[L\Big(h\Big(\boldsymbol{X} ; \bm\omega, \bm{\theta}\Big), Y\Big)\Big],
\end{equation}
where $\Theta$ is the hyperparameter search space and $L:\mathcal{Y}\times\mathcal{Y}\rightarrow\mathbb{R}^{+}$ is a continuous loss function which satisfies the Lipschitz condition. $f_T(\bm\theta)$ is the target risk function, which is defined as the risk function over the target distribution. $h\left(\bm{x};\bm{\omega},\bm{\theta}\right):\mathcal{X}\rightarrow\mathcal{Y}$ is a parametric model which predicts the output value using the feature instance $\bm x$ with the model parameter $\bm\omega$ when the hyperparameter is $\bm\theta\in\Theta$.

In a standard hyperparameter optimization setting,~see Ref.~\refcite{NIPS2011_86e8f7ab} and \refcite{Feurer2019}, the output values of the target dataset are available. Let $D_T^{op}=\{(\bm x_i,y_i)\}_{i=1}^{n_T^{op}}$ be the i.i.d. target dataset in this situation. Then, we can split $D_T^{op}$ to two disjoint subsets $D_T^{optr}\cup D_T^{opval}$. For a given hyperparameter $\bm\theta\in\Theta$, we estimate the target risk in (\ref{01}) by the following empirical mean:
\begin{equation*}\label{00}
\hat{f}_T(\bm{\theta};D_T^{opval})=\frac{1}{|D_T^{opval}|}\sum_{(\bm{x}_i,y_i)\in D_T^{opval}}L\Big(h\Big(\boldsymbol{x}_i ; \hat{\bm\omega}(\boldsymbol{\theta}), \boldsymbol{\theta}\Big), y_i\Big),
\end{equation*}
where $|D|$ denotes the sample size of the dataset $D$. The model parameter is estimated on $D_T^{optr}$ by
\begin{equation*}
\hat{\bm{\omega}}(\bm{\theta})=\underset{\bm{\omega}\in\Omega}{\arg\min}\frac{1}{|D_T^{optr}|}\sum_{(\bm x_i,y_i)\in D_T^{optr}}L\Big(h\Big(\bm{x}_i;\bm{\omega},\bm{\theta}\Big),y_i\Big),
\end{equation*}
where $\Omega$ is the model parameter space. Then, we can obtain the estimate of $\bm\theta^\star$ by minimizing $\hat{f}_T(\bm\theta;D_T^{opval})$. 

In contrast, in the FedCS setting of our paper, the output values of the target data are completely missing. Instead, we can utilize multiple source datasets whose output values are available, but there are covariate shifts between the sources and target. For covariate shift adaptation, the natural idea is to use importance sampling to approximate the target risk,~see Ref.~\refcite{2015Efficient} and \refcite{2007Covariate}.

\subsubsection{FedIWE}

Since every source does not allow the data to leave its local area or be disclosed, for a given hyperparameter $\bm{\theta}\in\Theta$, let the $j$th source learns locally to obtain the estimate of the model parameter $\hat{\bm\omega}_j(\bm\theta)$. Although all sources use the same hyperparameter $\bm\theta$, $\{\hat{\bm\omega}_j(\bm\theta)\}_{j=1}^{K}$ will be different because the source datasets are different.

Each source uses importance sampling in local learning, which yields a density ratio. We define the following density ratio.
\begin{definition}
For any $(\bm x,y)\in\mathcal{X}\times\mathcal{Y}$ with a positive source density $p_j(\bm x,y)>0$, the density ratio between the target and $j$th source is
$$r_j(\bm{x})=\frac{p_T(\bm{x},y)}{p_j(\bm{x},y)}=\frac{p_T(\bm{x})}{p_j(\bm{x})},j=1,2,\cdots,K.$$
The equalities are derived from Assumption \ref{assu02}.
\end{definition}
In the FedCS setting, according to Assumption \ref{assu01}, we can derive that $r_j(\bm x)\in [0,C]$ for a positive constant $C$.

Since density ratio estimation, training the model and selecting optimal hyperparameters are involved in our optimization process, we can split $D_j$ to three disjoint subsets $D_j^{de} \cup D_j^{tr}\cup D_j^{val}$, where 
$$D_j^{de}=\Big\{(\bm{x}_{j,i}^{de},y_{j,i}^{de})\Big\}_{i=1}^{n_{j}^{de}}, D_j^{tr}=\Big\{(\bm{x}_{j,i}^{tr},y_{j,i}^{tr})\Big\}_{i=1}^{n_{j}^{tr}}, D_j^{val}=\Big\{(\bm{x}_{j,i}^{val},y_{j,i}^{val})\Big\}_{i=1}^{n_j^{val}}.$$
Let $n_j=n_j^{de}+n_j^{tr}+n_j^{val}$, $n^{tr}=\sum_{j=1}^K n_j^{tr}$ and $n^{val}=\sum_{j=1}^K n_j^{val}$.

For a given hyperparameter $\bm\theta\in\Theta$, at the $j$th source, empirical estimate of the target risk can be adjusted as
\begin{equation}\label{03}
\hat f_{IW}(\bm{\theta};D_j)=\frac{1}{n_j^{val}}\sum_{i=1}^{n_j^{val}}\hat{r}_j(\bm{x}_{j,i}^{val})L\Big(h\Big(\bm{x}_{j,i}^{val};\hat{\bm{\omega}}_j(\bm{\theta}),\bm{\theta}\Big),y_{j,i}^{val}\Big),
\end{equation}

where the estimate of the density ratio $\hat r(\cdot)$ can be obtained on $D_j^{de}$ and $D_T$ using the unconstrained least-squares importance fitting (uLSIF) method,~see Ref.~\refcite{2009A} and \refcite{Yamada2013Relative}. The model parameter is estimated on $D_j^{tr}$ as
\begin{equation}\label{02}
\hat{\bm{\omega}}_j(\bm{\theta})=\underset{\bm{\omega}\in\Omega}{\arg\min}\frac{1}{n_j^{tr}}\sum_{i=1}^{n_j^{tr}}L\Big(h\Big(\bm{x}_{j,i}^{tr};\bm{\omega},\bm{\theta}\Big),y_{j,i}^{tr}\Big),
\end{equation}
 and the ideal model parameter is
\begin{equation*}
{\bm{\omega}}_j(\bm{\theta})=\underset{\bm{\omega}\in\Omega}{\arg\min}\mathbb{E}_{(\bm x,y)\sim p_j(\bm x,y)}\Big[L\Big(h\Big(\bm x;\bm{\omega},\bm{\theta}\Big),y\Big)\Big].
\end{equation*}

We demonstrate that (\ref{03}) is asymptotically unbiased for $f_T(\bm\theta)$ and has an asymptotic variance as follows. The proof is provided in the appendix.

\begin{theorem}\label{thm01}
For a given hyperparameter $\bm\theta\in\Theta$, suppose $h$ is continuous for $\bm\omega$, we have
\begin{romanlist}[(ii)]
\item $\lim\limits_{n_j^{de},n_j^{tr},n_T\to\infty}\mathbb{E}\Big[\hat{f}_{IW}(\bm\theta;D_j)\Big]=f_T(\bm\theta);$
\item $\lim\limits_{n_j^{de},n_j^{tr},\atop n_T\to\infty}\mathbb{V}\Big[\hat{f}_{IW}(\bm\theta;D_j)\Big]=\mathbb{E}_{(\bm{x},y)\sim p_j(\bm x,y)}\Big[r_j(\bm{x})L\Big(h\Big(\bm{x};\bm{\omega}_j(\bm{\theta}),\bm{\theta}\Big),y\Big)\Big]^2-f^2_T(\bm\theta).$

\end{romanlist}
\end{theorem}

Because each source learns a parametric model locally, their learning process does not involve other source data, satisfying the privacy protection in the FedCS setting. Then, we propose FedIWE of the target risk as
\begin{equation}\label{04}
\hat{f}_{FedIW}\Big(\bm\theta;\{D_j\}_{j=1}^K\Big)=\frac{1}{n^{val}}\sum_{j=1}^K n_j^{val}\hat{f}_{IW}(\bm\theta;D_j).
\end{equation}
Follow Theorem \ref{thm01} and the linear property of the expectation operation, we can derive that FedIWE is asymptotically unbiased for the target risk for any given $\bm\theta\in\Theta$, i.e.,
\begin{equation*}
\lim\limits_{\forall j,n_j^{de},n_j^{tr}, n_T\to\infty}\mathbb{E}\Big[\hat{f}_{FedIW}(\bm\theta;\{D_j\}_{j=1}^K)\Big]=f_T(\bm\theta).
\end{equation*}

Then, the estimate of $\bm\theta^\star$ is obtained by $$\hat{\bm\theta}_{FedIW}=\underset{\bm\theta\in\Theta}{\arg\min}\hat{f}_{FedIW}\Big(\bm\theta;\{D_j\}_{j=1}^K\Big),$$
and the corresponding estimates of the model parameters are $\{\hat{\bm\omega}_j(\hat{\bm\theta}_{FedIW})\}_{j=1}^K$. 

However, (\ref{03}) is unstable,~see Ref.~\refcite{2007Covariate}. From Theorem \ref{thm01}, when the $j$th source has a distribution dissimilar to that of the target, i.e., $r_j(\bm x),\bm x\in\mathcal{X}$ can be large, the variance of (\ref{03}) will become large as $n_j^{de},n_j^{tr},n_T\to\infty$. However, the estimate with large variance may not be accurate in the long run. 

\subsubsection{FedDAE}

Lemieux\cite{Lemieux2017} shows that the control variates method can reduce variance, so we adjust (\ref{03}) as follows, for a given hyperparameter $\bm\theta\in\Theta$,
\begin{equation}\label{07}
\hat f_{CV}(\bm{\theta};D_j)=\frac{1}{n_j^{val}}\sum_{i=1}^{n_j^{val}}\Big[\hat{r}_j(\bm{x}_{j,i}^{val})L\Big(h\Big(\bm{x}_{j,i}^{val};\hat{\bm{\omega}}_j(\bm{\theta}),\bm{\theta}\Big),y_{j,i}^{val}\Big)+\hat{\eta}_j(\bm\theta)\cdot\Big(\hat r_j(x_{j,i}^{val})-1\Big)\Big],
\end{equation}
where let $\bm x_j^{val}=(\bm x_{j,1}^{val},\cdots,\bm x_{j,n_j^{val}}^{val})$ and $y_j^{val}=(y_{j,1}^{val},\cdots,y_{j,n_j^{val}}^{val})$, 
\begin{equation*}
\hat{\eta}_{j}(\bm\theta)=-\frac{\widehat{\operatorname{Cov}}\Big[\hat{r}_{j}(\bm{x}_j^{val})\cdot L\Big(h\big(\bm{x}_j^{val};\hat{\bm{\omega}}_j(\bm{\theta}),\bm{\theta}\big),y_j^{val}\Big),\hat{r}_{j}(\bm{x}_j^{val})\Big]}{\widehat{\operatorname{Var}}\Big[\hat{r}_{j}(\bm{x}_j^{val})\Big]}, j=1,2,\cdots,K.
\end{equation*}
For $ j=1,2,\cdots,K$, we define
\begin{equation*}
\eta_{j}(\bm\theta)=-\frac{\operatorname{Cov}_{(\bm x,y)\sim p_j(\bm x,y)}\Big[r_{j}(\bm x)\cdot L\Big(h\Big(\bm x;\bm{\omega}_j(\bm{\theta}),\bm{\theta}\Big),y\Big),r_{j}(\bm x)\Big]}{\operatorname{Var}_{(\bm x,y)\sim p_j(\bm x,y)}\Big[r_{j}(\bm x)\Big]}.
\end{equation*}
By Theorem \ref{thm01} and the control variates method\cite{Lemieux2017}, we can obtain the following corollary and the proof is provided in the appendix.

\begin{corollary}\label{cor01}
For a given hyperparameter value $\bm\theta\in\Theta$, under the same conditions in Theorem \ref{thm01}, suppose $\operatorname{Var}_{\bm x\sim p_j(\bm x)}\left[r_j(\bm x)\right]\neq 0,j=1,2\cdots,K$. Then 
\begin{romanlist}[(ii)]
\item $\lim\limits_{n_j^{de},n_j^{tr},n_T\to\infty}\mathbb{E}\Big[\hat{f}_{CV}(\bm\theta;D_j)\Big]=f_T(\bm\theta);$
\item $\lim\limits_{n_j^{de},n_j^{tr},n_T\to\infty}\mathbb{V}\Big[\hat{f}_{CV}(\bm\theta;D_j)\Big]\leq\lim\limits_{n_j^{de},n_j^{tr},n_T\to\infty}\mathbb{V}\Big[\hat{f}_{IW}(\bm\theta;D_j)\Big].$

\end{romanlist}

\end{corollary}

It shows that $\hat{f}_{CV}(\bm\theta;D_j)$ is more stable than $\hat{f}_{IW}(\bm\theta;D_j)$ as $n_j^{de},n_j^{tr},n_T\to\infty$, so we further adjust the empirical estimate of the target risk as $\hat{f}_{CV}(\bm\theta;D_j)$, $j=1,2,\cdots,K$. Then we define a class of asymptotic unbiased estimates of the target risk.

\begin{definition}
For a given hyperparameter $\bm\theta\in\Theta$, we define the federated $\bm\lambda$ estimates (FedLE) for the target risk as
$$\hat f_{\bm\lambda}\Big(\bm{\theta};\{D_j\}_{j=1}^K\Big)=\sum_{j=1}
^K\lambda_j n_j^{val}\cdot\hat{f}_{CV}(\bm\theta;D_j),$$
where $\bm\lambda=\{\lambda_1,\cdots,\lambda_K\}$ is any set of weights for sources that satisfies $\lambda_j\geq 0$, $j=1,2,\cdots,K$, and $\sum_{j=1}^K\lambda_j n_j^{val}=1$.
\end{definition}

As shown in~Ref.~\refcite{nomura2021efficient}, we can construct an ingenious way to integrate $\{\hat f_{CV}(\bm\theta;D_j)\}_{j=1}^K$. As a preliminary, we first define a divergence measure, which quantifies the similarity between the source and target.

\begin{definition}\label{def03}
(Divergence Measure) For a given hyperparameter $\bm\theta\in\Theta$, the divergence measure between the $j$th source and target is defined as
\begin{equation}\label{08}
\begin{aligned}
\operatorname{Div}_j\left(\bm\theta\right)=&\mathbb{E}_{(\bm{X}, Y) \sim p_j(\bm x,y)}\Big[r_j(\bm{X}) \cdot L\Big(h\Big(\bm{X};\bm{\omega}_j(\bm{\theta}),\bm{\theta}\Big), Y\Big)+\eta_j(\bm\theta)\cdot\Big(r_j(\bm{X})-1\Big)\Big]^2\\
&-f^2_{T}(\bm{\theta}), j=1,2,\cdots,K.
\end{aligned}
\end{equation}
\end{definition}

This divergence measure is large when the corresponding
source distribution deviates significantly from the target distribution. Here, we estimate (\ref{08}) on the source data as
\begin{equation*}
\begin{aligned}
\widehat{\operatorname{Div}}_j\left(\bm\theta\right)=\frac{1}{n_j^{val}}\sum_{i=1}^{n_j^{val}} \Big[\hat{r}_j(\bm{x}_{j,i}^{val}) \cdot L\Big(h\Big(\bm{x}_{j,i}^{val};\hat{\bm{\omega}}_j(\bm{\theta}),\bm{\theta}\Big), y_{j,i}^{val}\Big)+\hat{\eta}_j(\bm\theta)\cdot\Big(\hat{r}_j(\bm{x}_{j,i}^{val})-1\Big)\Big]^2\\-\Big[\frac{1}{n_j^{val}}\sum_{i=1}^{n_j^{val}}\Big(\hat{r}_j(\bm{x}_{j,i}^{val}) \cdot L\Big(h\Big(\bm{x}_{j,i}^{val};\hat{\bm{\omega}}_j(\bm{\theta}),\bm{\theta}\Big), y_{j,i}^{val}\Big)+\hat{\eta}_j(\bm\theta)\cdot\Big(\hat{r}_j(\bm{x}_{j,i}^{val})-1\Big)\Big)\Big]^2.
\end{aligned}
\end{equation*}

Based on the divergence measure, we propose FedDAE of the target risk as
\begin{equation}\label{09}
\hat f_{FedDA}\Big(\bm{\theta};\{D_j\}_{j=1}^K\Big)=\sum_{j=1}
^K\hat\lambda_j(\bm\theta) n_j^{val}\cdot\hat{f}_{CV}(\bm\theta;D_j),
\end{equation}
where the weight of the $j$th source is defined as
\begin{equation}\label{10}
\hat\lambda_j(\bm\theta)=\Bigg(\widehat{\operatorname{Div}}_j\left(\bm\theta\right) \sum_{j=1}^K \frac{n_j^{val}}{\widehat{\operatorname{Div}}_j\left(\bm\theta\right)}\Bigg)^{-1}.
\end{equation}
Note that $\hat\lambda_j(\bm\theta)\geq0$, $j=1,2,\cdots,K$, and 
\begin{equation*}
\begin{aligned}
\sum_{j=1}^K\hat\lambda_j(\bm\theta)n_j^{val}=&\sum_{j=1}^K\Bigg(\widehat{\operatorname{Div}}_j\left(\bm\theta\right) \sum_{j=1}^K \frac{n_j^{val}}{\widehat{\operatorname{Div}}_j\left(\bm\theta\right)}\Bigg)^{-1}\cdot n_j^{val}\\
=&\sum_{j=1}^K\Bigg(\sum_{j=1}^K \frac{n_j^{val}}{\widehat{\operatorname{Div}}_j\left(\bm\theta\right)}\Bigg)^{-1}\Bigg(\frac{n_j^{val}}{\widehat{\operatorname{Div}}_j\left(\bm\theta\right)}\Bigg)\\
=&\Bigg(\sum_{j=1}^K \frac{n_j^{val}}{\widehat{\operatorname{Div}}_j\left(\bm\theta\right)}\Bigg)^{-1}\Bigg(\sum_{j=1}^K \frac{n_j^{val}}{\widehat{\operatorname{Div}}_j\left(\bm\theta\right)}\Bigg)\\
=&1.
\end{aligned}
\end{equation*}
For $j=1,2,\cdots,K$, we define 
$$\lambda_{j}(\bm\theta)=\Bigg(\operatorname{Div}_j\left(\bm\theta\right) \sum_{j=1}^K \frac{n_j^{val}}{\operatorname{Div}_j\left(\bm\theta\right)}\Bigg)^{-1}.$$
We investigate the asymptotic property of FedDAE below.
\begin{theorem}\label{cor02}
For a given hyperparameter $\bm\theta\in\Theta$, under the same conditions in Corollary \ref{cor01}, suppose $\eta_j(\bm\theta)$ is uniform bounded, $j=1,2,\cdots,K$. Then, for any given set of weights for sources $\bm\lambda$, 

\begin{romanlist}[(ii)]
\item $\lim\limits_{\forall j,n_j^{de},n_j^{tr},n_T\to\infty}\mathbb{E}\Big[\hat f_{\bm\lambda}\Big(\bm{\theta};\{D_j\}_{j=1}^K\Big)\Big]=f_T(\bm\theta);$

Especially, $\lim\limits_{\forall j,n_j^{de},n_j^{tr},n_T\to\infty}\mathbb{E}\Big[\hat f_{FedDA}\Big(\bm{\theta};\{D_j\}_{j=1}^K\Big)\Big]=f_T(\bm\theta).$
\item $\lim\limits_{\substack{\forall j,n_j^{de},n_j^{tr},\atop n_j^{val},n_T\to\infty}}\mathbb{V}\Big[\hat f_{FedDA}\Big(\bm{\theta};\{D_j\}_{j=1}^K\Big)\Big]\leq\lim\limits_{\substack{\forall j,n_j^{de},n_j^{tr},\atop n_j^{val},n_T\to\infty}}\mathbb{V}\Big[\hat f_{\bm\lambda}\Big(\bm{\theta};\{D_j\}_{j=1}^K\Big)\Big].$

\end{romanlist}

\end{theorem}

The inequality in Corollary \ref{cor02} demonstrates that FedDAE has the smallest asymptotic variance in FedLE, that is a class of asymptotically unbiased estimates of the target risk. Follow Corollary \ref{cor01} and Corollary \ref{cor02}, we can easily obtain the follow inequality and provide the proof in the appendix.
\begin{corollary}\label{cor03}
For a given hyperparameter $\bm\theta\in\Theta$, under the same conditions in Corollary \ref{cor02}, then
\begin{equation*}
\lim\limits_{\substack{\forall j,n_j^{de},n_j^{tr},\atop n_j^{val},n_T\to\infty}}\mathbb{V}\Big[\hat f_{FedDA}\Big(\bm{\theta};\{D_j\}_{j=1}^K\Big)\Big]\leq\lim\limits_{\substack{\forall j,n_j^{de},n_j^{tr},\atop n_j^{val},n_T\to\infty}}\mathbb{V}\Big[\hat f_{FedIW}\Big(\bm{\theta};\{D_j\}_{j=1}^K\Big)\Big].
\end{equation*}
\end{corollary}

It shows that the asymptotic variance of FedDAE is smaller than that of FedIW, so we futher adjust the estimate of the target risk as FedDAE. Then, we estimate $\bm\theta^\star$ by
\begin{equation*}\label{11}
\hat{\bm\theta}_{FedDA}=\underset{\bm\theta\in\Theta}{\arg\min}\hat{f}_{FedDA}\Big(\bm\theta;\{D_j\}_{j=1}^K\Big).
\end{equation*}
The corresponding estimates of the model parameters and the weights of the sources are $\{\hat{\bm\omega}_j(\hat{\bm\theta}_{FedDA})\}_{j=1}^K$ and $\{\hat{\lambda}_j(\hat{\bm\theta}_{FedDA})\}_{j=1}^K$, respectively. Thus, we obtain the model learned locally by the $j$th source, i.e., $h(\cdot;\hat{\bm\omega}_j(\hat{\bm\theta}_{FedDA}),\hat{\bm\theta}_{FedDA})$, $j=1,2,\cdots,K$. Next we have to consider how to use all source models for the execution of the target task.

\subsection{FedCSA algorithm}

As shown in~Ref.~\refcite{peng2019federated}, we can integrate these models as a weighted model. Choosing the reasonable weights $\alpha_1,\cdots,\alpha_K$ for the source models can prevent negative transfer\cite{2010A} from occurring and the empirical research shows that if the distribution difference between two domains is too significant, the rough transfer may affect the implementation of the target task,~see Ref.~\refcite{smith2001transfer}. 

We define the weight of the $j$th source model based on the weight of the $j$th source in FedDAE, and use the following weighed model
\begin{equation}\label{12}
\hat h_T(\cdot)=\sum_{j=1}^K\alpha_j h(\cdot;\hat{\bm\omega}_j(\hat{\bm\theta}_{FedDA}),\hat{\bm\theta}_{FedDA}),
\end{equation}
where $\alpha_j=\hat{\lambda}_j(\hat{\bm\theta}_{FedDA})n_j^{val}$ is the weight of the $j$th source model. Note that $\alpha_j\geq 0,j=1,2\cdots,K$ and $\sum_{j=1}^K\alpha_j=1$. The experimental results in the next two sections verify the weighted model (\ref{12}) works well. Refer to Peng \emph{et al}.\cite{peng2019federated} and Blitzer \emph{et al}.\cite{blitzer2007learning}, we have the following lemma.

\begin{lemma}\label{lem01}
Let $\epsilon_T(h)$ and $\hat{\epsilon}_S(h)$ be the error of the target task and the empirical error on the mixture of source datasets with size $n=\sum_{j=1}^K n_j$. Then, $\forall\alpha_j>0,\sum_{j=1}^K\alpha_j=1$, with probability at least $1-\delta$ over the choice of samples, for a given hyperparameter $\bm\theta\in\Theta$,
\begin{equation}\label{13}
\begin{aligned}
\epsilon_T(\hat h_T(\cdot))\leq&\hat{\epsilon}_S\Big(\sum_{j=1}^K\alpha_jh(\cdot;\hat{\bm\omega}_j(\bm\theta),\bm\theta)\Big)+\sum_{j=1}^K\alpha_j\Big(\frac{1}{2}\widehat{d}_{\mathcal{H} \Delta \mathcal{H}}\left(D_j, D_T\right)+t_i\Big)\\
&+4\sqrt{\frac{2  \log (2 n)+\log (4 / \delta)}{n}},
\end{aligned}
\end{equation}
where $t_i$ is the error of the optimal model on the mixture of $D_j$ and $D_T$, and $\widehat{d}_{\mathcal{H} \Delta \mathcal{H}}\left(D_j, D_T\right)$ denotes the divergence involving $\mathcal{H} \Delta \mathcal{H}$ discrepancy\cite{ben2010theory} between the $j$th source and target. 
\end{lemma}
The details of the proof are provided in Peng \emph{et al}.\cite{peng2019federated} and Blitzer \emph{et al}.\cite{blitzer2007learning}. Especially, (\ref{13}) holds for $\hat{\bm\theta}$.

We propose the FedCSA algorithm which use FedDAE for the target risk. This algorithm can also use other estimates. Later we will compare FedIWE and FedDAE. For a given hyperparameter $\bm\theta\in\Theta$, we write a shorthand $\hat f(\bm\theta)=\sum_{j=1}^{K}\hat\lambda_j(\bm\theta) n_j^{val}\hat f_{j}(\bm\theta)$. For FedDAE, $\hat f$ is $\hat f_{FedDA}(\bm\theta;\{D_j\}_{j=1}^K)$ and $\hat f_j$ is $\hat f_{CV}(\bm\theta;D_j)$. For FedIWE, $\hat f$ is $\hat f_{FedIW}(\bm\theta;\{D_j\}_{j=1}^K)$, $\hat f_j$ is $\hat f_{IW}(\bm\theta;D_j)$ and $\hat\lambda_j(\bm\theta)$ is $1/n_j^{val}$.

For any given hyperparameter $\bm{\theta}\in\Theta$, let the $j$th source learns a parametric model locally, then transmits $\hat{f}_j(\bm\theta)$, $\hat{\bm\omega}_j(\bm\theta)$, $\hat{\lambda}_j(\bm\theta)$ and $n_j^{val}$ to the target. The target estimate $\bm\theta^\star$ by
$$\hat{\bm{\theta}} =\underset{\bm{\theta}\in\Theta}{\arg \min}\hat{f}(\bm{\theta})=\underset{\bm{\theta}\in\Theta}{\arg\min}\sum_{j=1}^K\hat\lambda_j(\bm\theta)n_j^{val}\hat f_j(\bm\theta).$$
The corresponding estimate of the model parameter and the source weight are $\hat{\bm\omega}_j(\hat{\bm\theta})$ and $\hat{\lambda}_j(\hat{\bm\theta})$. Finally, the target obtains the weighted model as $$\hat h_T(\cdot)=\sum_{j=1}^{K}\hat\alpha_j h\Big(\cdot;\hat{\bm{\omega}}_j(\hat{\bm{\theta}});\hat{\bm{\theta}}\Big),\hat\alpha_j=\hat{\lambda}_j(\hat{\bm\theta})n_j^{val},j=1,2,\cdots,K.$$

\begin{algorithm}
\renewcommand{\algorithmicrequire}{\textbf{Input:}}
\renewcommand{\algorithmicensure}{\textbf{Output:}}
\caption{FedCSA algorithm\label{alg1}}
\begin{algorithmic}[1]

\REQUIRE Target data $D_T=\{(\bm{x}_{T,i})\}_{i=1}^{n_T}$; Source data $D_j=\{(\bm{x}_{j,i},y_{j,i})\}_{i=1}^{n_j},j=1,2,\cdots,K$; Hyperparameter search space $\Theta$; Parametric model $h$; Estimate of the target risk $\hat f(\bm\theta)=\sum_{j=1}^K\hat\lambda_j(\bm\theta)n_j^{val}\hat f_j(\bm\theta)$.

\ENSURE The prediction model of the target task, $\hat h_T(\cdot)$.

\STATE {The target transmits $D_T$ to each source separately.}

\FOR {$j=1$ to $K$}

\STATE {The $j$th \textbf{source}:
\begin{enumerate}
\item Split $D_j$ as three disjoint folds: $D_j^{de},D_j^{tr},D_j^{val}$;

\item Estimate density ratio by uLSIF\cite{2009A} on $D_T$ and $D_j^{de}$, record as $\hat{r}_j(\cdot)$ ;

\item Learn locally on $D_j^{tr}$ and estimate the model parameter as $\hat{\bm{\omega}}_j(\bm{\theta})$ by (\ref{02});

\item Calculate $\hat{f}_j(\bm\theta)$ and $\hat{\lambda}_j(\bm\theta)$ on $D_j^{val}$ by (\ref{03}) or (\ref{07}) and (\ref{10});

\item Transmit $\hat{f}_j(\bm\theta)$, $\hat{\bm{\omega}}_j(\bm{\theta})$, $\hat{\lambda}_j(\bm\theta)$ and $n_j^{val}$ to the target.
\end{enumerate}
}
\ENDFOR

\STATE {The \textbf{target}:

\begin{enumerate}
\item Obtain $\hat f(\bm\theta)=\sum_{j=1}^K\hat{\lambda}_j(\bm\theta)n_j^{val}\hat{f}_j(\bm{\theta});$
\item Estimate $\hat{\bm{\theta}} =\underset{\bm{\theta}\in\Theta}{\arg \min}\hat f(\bm{\theta});$

\item Determine $\hat{\bm\omega}_j(\hat{\bm\theta})$ and $\hat{\lambda}_j(\hat{\bm\theta})$, $j=1,2,\cdots,K$;
\item Derive $\hat\alpha_j=\hat{\lambda}_j(\hat{\bm\theta})n_j^{val}$, $j=1,2,\cdots,K$.
\end{enumerate}
}
\STATE {\textbf{Return} $\hat h_T(\cdot)=\sum_{j=1}^K\hat\alpha_j h\Big(\cdot;\hat{\bm{\omega}}_j(\hat{\bm{\theta}});\hat{\bm{\theta}}\Big)$.}

\end{algorithmic}
\end{algorithm}



\section{Simulation Studies}\label{SecSimul}

In this section, we perform some simulations to illustrate the performance of our proposed method. We consider two cases, one is to study the performance of the method under different sample sizes, and the other is to study the performance of the method under different covariate shifts between the sources and target. The data generation progress is as follows:

\textbf{Case 1:} The input variable $\bm{x}=(\bm{x}_1,\bm{x}_2,\cdots,\bm{x}_{10})$ is 10-dimensional. There are two sources, their distributions are $N(\bm{1}_{10},3\times\bm{I}_{10})$ and $N(5\times\bm{1}_{10},0.5\times\bm{I}_{10})$. $\bm{1}_{10}$ is the 10-dimensional vector with all ones and $\bm{I}_{10}$ is the 10-dimensional identity matrix. The sample sizes of two sources are $n_1$ and $n_2$. The distribution of the target is $N(\bm{0}_{10},\bm{I}_{10})$,  $\bm{0}_{10}$ is the 10-dimensional vector with all zeros. The sample size of the target is $n_T$. The output variate $y$ is 1-dimensional. The conditional output distribution is $y|\bm{x}\sim N(\tilde{\bm{x}},1)$, $\tilde{\bm{x}}=\frac{1}{10}\sum^{10}_{i=1}\bm{x}_i$. 

\textbf{Case 2:} Two sources, the distributions are $N(c\times\bm{1}_{10},\bm{I}_{10})$ and $N((c+1)\times\bm{1}_{10},\bm{I}_{10})$. $c\in\{1,1.5,2,2.5,3,3.5,4,4.5,5\}$. Their sample sizes are 50 and 40. The distribution of the target domain is $N(\bm{0}_{10},\bm{I}_{10})$, and its sample size is 20. The others are the same as that of Case 1.

We use ridge regression model and the square loss function: $L(\hat y,y)=(\hat y-y)^2$. The hyperparameter is the regularization parameter $\lambda$, and the hyperparameter search space is $\Theta=[0,1]$. We report the prediction error with mean absolute error: $\epsilon=\frac{1}{n}\sum_{i=1}^n |\hat y_i-y_i|$. We compare four methods and FedDA represents our proposed method:
\begin{romanlist}[(ii)] 
\item FedIW: use (\ref{04}) as $\hat f(\bm\theta)$ in Algorithm \ref{alg1}. Especially, $\hat\lambda_j=1/n_j^{val}$ and $\alpha_j=n_j^{val}/n^{val}$, $j=1,2,\cdots,K$. They don't rely on the hyperparameter $\bm\theta$.
\item FedDA: use (\ref{08}) as $\hat f(\bm\theta)$ in Algorithm \ref{alg1}. 
\item Naive: select the parametric model directly without considering covariate shifts between the source and target domains.
\item Reference: use the target data with output values to learn a parametric model, that are not feasible in the FedCS setting and report its performance as a reference.
\end{romanlist}

\begin{figure}[h!t]
  \centering
  \subfigure[Comparing all methods]{
  \label{fig2a}
      \begin{minipage}[t]{0.45\linewidth}
      \centering
      \includegraphics[width=5cm]{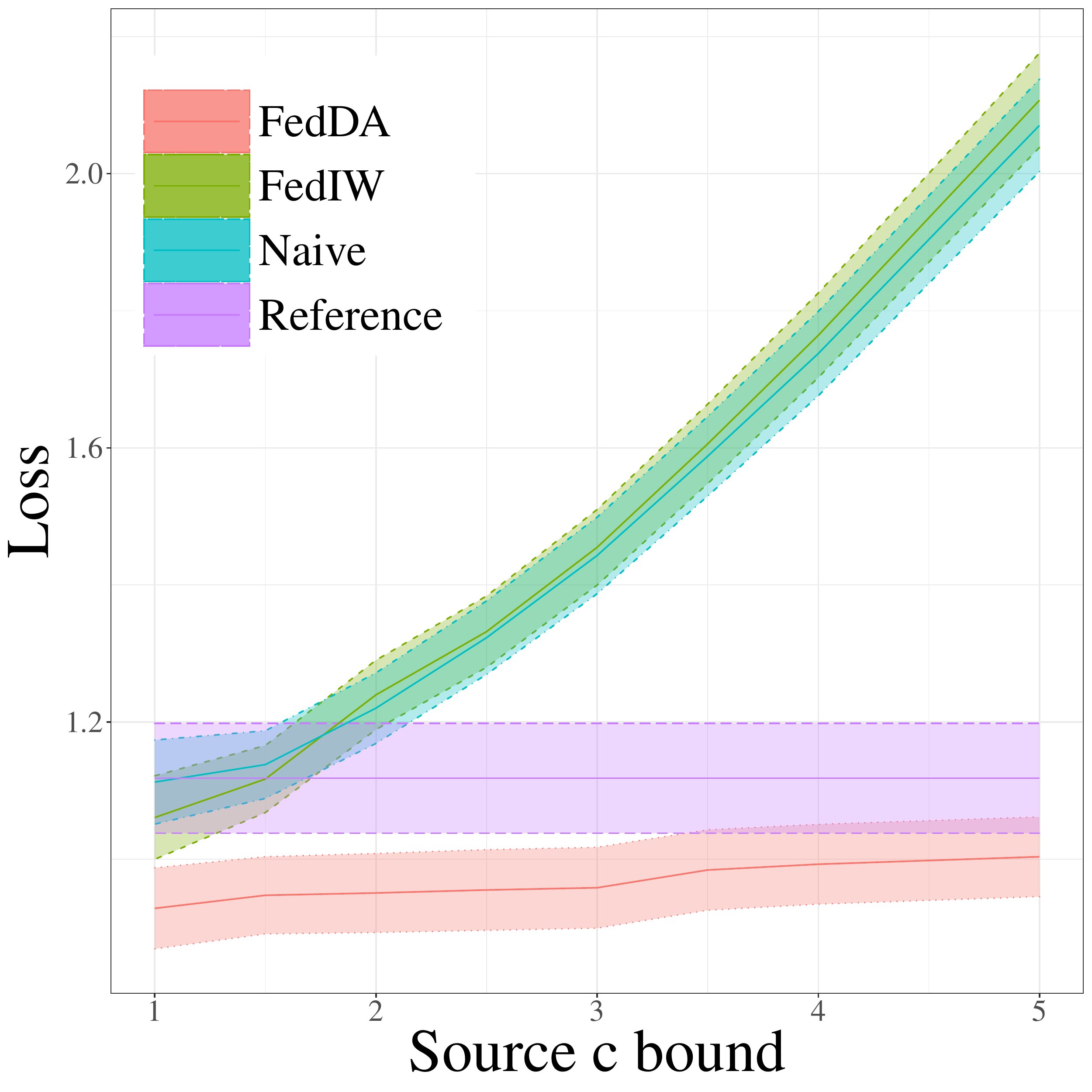}
   
      \end{minipage}
  }
  \subfigure[Comparing FedIW and FedDA]{
  \label{fig2b}
      \begin{minipage}[t]{0.45\linewidth}
      \centering
      \includegraphics[width=5cm]{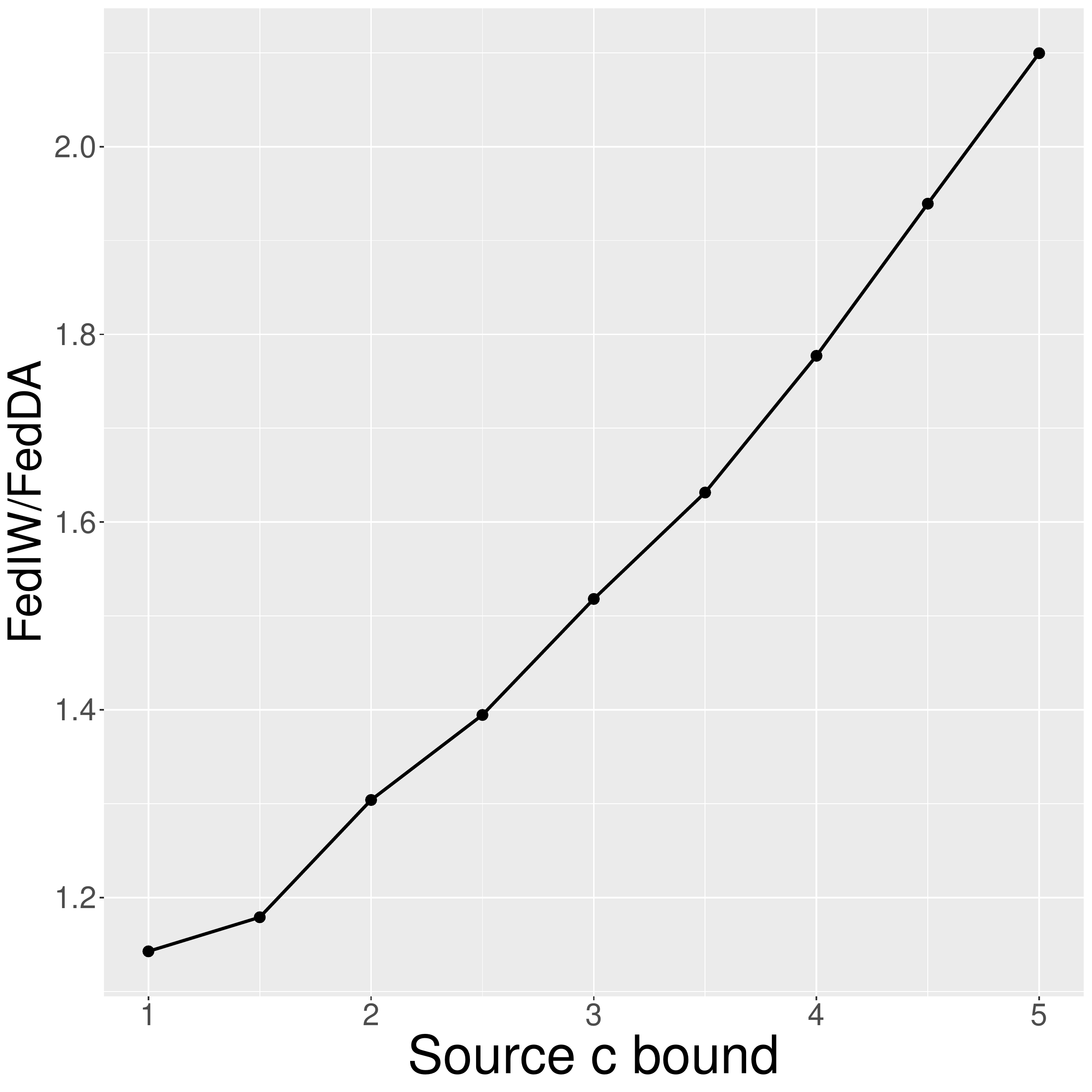}
     
      \end{minipage}
  }
  \caption{Performances of methods with different covariate shifts. The horizontal axis represents the parameters
of the source domains $c\in\{1,1.5,2,2.5,3,3.5,4,4.5,5\}$. (a) The vertical axis represents the mean of the
performance of each method. (b) The vertical axis represents the ratio of the mean performance of FedIW to FedDA}
  \label{fig2}
\end{figure}

In Case 1, we run 100 experiments with different random seeds for different sample sizes. The mean and the standard error of the 100 experimental results for every method are shown in Table \ref{table1}. In Case 2, for any $c\in\{1,1.5,2,2.5,3,3.5,4,4.5,5\}$, we run 100 experiments with different random seeds and the experimental results are shown in (Fig. \ref{fig2}). Fig. \ref{fig2a} shows the approximate 95\% confidence interval for the average performance of each method. Fig. \ref{fig2b} shows the ratio of the average performance of FedIW and FedDA.

\begin{table}[h!t]
\tbl{Performances (Mean$\pm$StdErr) of methods with diﬀerent sample sizes\label{table1}}
{\begin{tabular}{@{}c c c|c c c c@{}}
\toprule
$n_T$ &$n_1$ &$n_2$  &FedIW &FedDA &Naive &Reference  \\
\colrule
\multirow{5}{1cm}{\centering 20}
 &30 &20 &1.3548$\pm$0.3994 &\textbf{0.8649$\pm$0.3862} &1.6049$\pm$0.4620 &1.1222$\pm$0.2961\\
 &40 &30 &5.5673$\pm$0.9708 &\textbf{1.0530$\pm$0.2413} &4.6059$\pm$0.4027 &1.5219$\pm$0.3845\\
 &50 &40 &1.7531$\pm$0.4173 &\textbf{0.5635$\pm$0.1646} &1.7646$\pm$0.4096 &0.9178$\pm$0.2866\\
 &60 &50 &2.4793$\pm$0.4965 &\textbf{0.8353$\pm$0.2340} &2.5488$\pm$0.4832 &1.2125$\pm$0.2962\\
 &70 &60 &1.8884$\pm$0.4445 &\textbf{1.1029$\pm$0.2694} &1.9818$\pm$0.4622 &1.9644$\pm$0.4675\\
 \colrule
 \multirow{5}{1cm}{\centering 30}
 &40 &30 &3.3341$\pm$0.3237 &\textbf{0.7470$\pm$0.1815} &3.9530$\pm$0.3446 &0.8630$\pm$0.1847\\
 &50 &40 &1.7427$\pm$0.2686 &\textbf{0.8093$\pm$0.1530} &1.9016$\pm$0.2760 &1.2516$\pm$0.2745\\
 &60 &50 &1.9798$\pm$0.2976 &\textbf{0.8922$\pm$0.2017} &2.1204$\pm$0.3008 &1.1003$\pm$0.2598\\
 &70 &60 &4.1626$\pm$0.2907 &\textbf{0.6271$\pm$0.1230} &4.5362$\pm$0.2966 &0.7376$\pm$0.1413\\
 &80 &70 &3.0442$\pm$0.2679 &\textbf{0.8102$\pm$0.1511} &3.2368$\pm$0.2808 &1.0502$\pm$0.2951\\
 \colrule
 \multirow{5}{1cm}{\centering40}
 &50 &40 &1.8087$\pm$0.3199 &\textbf{0.7836$\pm$0.1462} &1.9628$\pm$0.3363 &0.8702$\pm$0.1836\\
 &60 &50 &5.2020$\pm$0.2335 &\textbf{0.8926$\pm$0.1479} &5.7482$\pm$0.2241 &0.8993$\pm$0.2119\\
 &70 &60 &4.9135$\pm$0.3448 &\textbf{0.9632$\pm$0.1878} &5.3687$\pm$0.3548 &1.1801$\pm$0.2110\\
 &80 &70 &1.5635$\pm$0.2199 &\textbf{0.7066$\pm$0.1217} &1.7052$\pm$0.2222 &0.7929$\pm$0.1563\\
 &90 &80 &1.9881$\pm$0.2415 &\textbf{0.8646$\pm$0.1760} &2.0885$\pm$0.2468 &0.9280$\pm$0.2259\\
 \colrule
 \multirow{5}{1cm}{\centering50}
 &60 &50 &2.8928$\pm$0.3238 &\textbf{0.7966$\pm$0.1375} &3.1662$\pm$0.3347 &0.8658$\pm$0.1386\\
 &70 &60 &1.4436$\pm$0.1979 &\textbf{0.6825$\pm$0.1176} &1.5184$\pm$0.2033 &0.8381$\pm$0.1441\\
 &80 &70 &2.2632$\pm$0.2607 &\textbf{0.6893$\pm$0.1185} &2.1614$\pm$0.2255 &0.8567$\pm$0.1229\\
 &90 &80 &2.2903$\pm$0.2459 &\textbf{0.7467$\pm$0.1314} &2.3971$\pm$0.2494 &0.8922$\pm$0.1599\\
 &100 &90 &0.9640$\pm$0.1457 &\textbf{0.7120$\pm$0.1282} &0.9897$\pm$0.1465 &0.8717$\pm$0.1335\\
\botrule
\end{tabular}}
\end{table} 

From the results of Case 1 in Table \ref{table1}, we can see that FedDA has the best mean performance with the smallest standard error in each situation. The results of Case 2 in Fig \ref{fig2} indicate that FedDA significantly outperforms the other methods for every value of $c$. Fig. \ref{fig2a} shows that the advantage of FedDA becomes more and more strengthened when the covariate shifts between the target and sources become larger. The curve in Fig. \ref{fig2b} shows that FedDA outperforms FedIW to an increasing degree as the covariate shifts between the target and sources become larger. FedDA significantly outperforms Naive because FedDAE can well adapt to the FedCS setting, while the Naive method does not. FedDA also outperforms FedIW because FedDAE has a smaller asymptotic variance than FedIWE. Compared with Reference, FedDA performs better because it can make full use of additional source samples. These results are consistent with our theoretical analysis.

\section{Real data Analysis}\label{SecReal}

We verify the performance of our proposed method on a real dataset, Parkinson's remote monitoring data.\cite{tsanas2009accurate} The dataset consists of a series of biomedical voice measurement data from 42 patients with early Parkinson's disease recruited to participate in a 6-month remote monitoring symptom progression test. These recordings were automatically recorded at the patients' home. Here, we regard each patient's home as a small hospital, and all the recordings of the patient are the data owned by this hospital. The columns of the dataset contain subject number, subject age, subject gender, time interval from baseline recruitment date, total UPDRS (Parkinson's disease score scale) score, and 16 biomedical voice measurements. Each row corresponds to one of the 5875 voice records of these people. The primary purpose of the dataset is to predict the total UPDRS score from the 16 voice measurements.

We choose one hospital as the target and the others as the sources. The patient disease data of different source hospitals can not be fused. We use two models to train the data: the weighted least squares with exponential weight and the ridge regression. The hyperparameters are the flattening parameter of the weight and the regularization parameter, respectively. The loss function is square loss.

Experimental process: Use the target data without output values and the source data with output values; Divide the original target data into the training set and test set according to 7:3; Train the model on the training set of the target data and the source data; Predict the disease scores of the test set of the target data; Calculate the mean absolute error of the prediction as to the performance of the method used; Repeat the above steps 100 times with different random seeds.

Report the mean, standard error and worst-case performances of the 100 experimental results of the weighted least squares and the ridge regression in Table \ref{table02} and Table \ref{table03}, respectively. The results indicate that FedDA is better than other methods in mean and worst-case with the smallest standard error.

\begin{table}[h!t]
\tbl{Weighted least squares on real data\label{table02}}
{\begin{tabular}{@{}c c c c@{}}  \toprule
Methods  &Mean  &Standard error  &Worst Case\\
\colrule
Naive &0.9003 &0.0753 &1.0862\\
FedIW &0.9519 &0.0871 &1.1573\\
FedDA  &\textbf{0.8578}  &\textbf{0.0617}  &\textbf{1.0034}\\
Reference &0.8834 &0.0657 &1.0524\\
\botrule
\end{tabular}}
\end{table}

\begin{table}[h!t]
\tbl{Ridge regression on real data\label{table03}}
{\begin{tabular}{@{}c c c c@{}}  \toprule
Methods  &Mean  &Standard error  &Worst Case\\
\colrule
Naive &0.8645 &0.0587 &1.0072\\
FedIW &0.8694 &0.0588 &1.0121\\
FedDA  &\textbf{0.8468}  &\textbf{0.0577}  &\textbf{0.9848}\\
Reference &0.8578 &0.0684 &1.0312\\
\botrule
\end{tabular}}
\end{table}

FedDA performs better than Naive because FedDAE can adapt to the FedCS setting between the hospitals. FedDA performs better than FedIW because FedDAE is stabler than FedIWE asymptotically. FedDA performs better than Reference because the samples of the source hospitals are more sufficient than that of the target hospital, and FedDA can make full use of existing data information.

\section{Conclusion}\label{SecCon}

This paper explores a new problem setting that extend the MS-CS setting under the framework of federated learning. The output values of the target data are completely missing, while the output values of the source data are available, and each source does not allow its data to leave its local area or be disclosed. To estimate the optimal hyperparameter, we first propose the federated importance weighting estimate of the target risk, which is asymptotically unbiased and can adapt to covariate shift. We further propose the federated domain adaptation estimate and show that it achieves the smaller asymptotic variance among a class of asymptotically unbiased estimates of the target risk. Then, we construct a weighted model by weighting all source models for the target task whose error can be bounded. Finally, we propose the federated covariate shift adaptation algorithm, a general and tractable hyperparameter optimization process. The experimental results indicate that our method is effective. However, our method does not take into account the nature of the under-sampled situation, which deserves further study in our future work.

\section*{Acknowledgments}
We acknowledge support from the National Natural Science Foundation of China (Grant Nos. 71873128, 12171451) .


\bibliographystyle{plain}
\bibliography{fedcsa}

\newpage
\appendix

\section{}

\textbf{A1. Lemma1}
Consider $\{f_n(y)\}$ is a sequence of functions
defined on the interval $[a,b]$. Suppose $f_n(y)$ converges uniformly to $f(y):[a,b]\to\mathbb{R}$ on $[a,b]$ as $n\to\infty$, then $\lim\limits_{n\to\infty}\underset{y\in[a,b]}{\min}f_n(y)=\underset{y\in[a,b]}{\min}f(y)$.

\textbf{Proof:} For any $\epsilon>0$, take $y_0\in[a,b]$ such that
\begin{equation}\label{A1}
    f(y_0)\leq \underset{y\in[a,b]}{\min}f(y)+\epsilon.
\end{equation}

then since
\begin{equation*}
    f(y_0)=\lim\limits_{n\to\infty}f_n(y_0)\geq\lim\limits_{n\to\infty}\underset{y\in[a,b]}{\min}f_n(y),
\end{equation*}

let $\epsilon\to 0$ we get
\begin{equation}\label{A2}
    \underset{y\in[a,b]}{\min}f(y)\geq\lim\limits_{n\to\infty}\underset{y\in[a,b]}{\min}f_n(y).
\end{equation}

Take a positive constant $N$ such that $\forall n\geq N$, $|f_n(y)-f(y)|\leq\epsilon$ for any $y\in[a,b]$. This gives that $\forall n\geq N$,
\begin{equation}\label{2ineq}
\begin{aligned}
   &-\epsilon\leq f_n(y)-f(y)\leq\epsilon,\\
   &-\epsilon\leq f_n(y_0)-f(y_0)\leq\epsilon.
   \end{aligned}
\end{equation}

From (\ref{2ineq}) we can obtain that
\begin{equation*}
\begin{aligned}
&-2\epsilon\leq f_n(y)-f(y)+f(y_0)-f_n(y_0)\leq 2\epsilon\\
&\implies f(y)-f_n(y)+f_n(y_0)\leq f(y_0)+2\epsilon.
\end{aligned}
\end{equation*}

Then
\begin{equation*}
\begin{aligned}
    \underset{y\in[a,b]}{\min}f(y)&= \underset{y\in[a,b]}{\min}[f(y)-f_n(y)+f_n(y)-f_n(y_0)+f_n(y_0)]\\
    &\leq \underset{y\in[a,b]}{\min}[f(y_0)+2\epsilon+f_n(y)-f_n(y_0)]\\
    &=\underset{y\in[a,b]}{\min}f_n(y)+f(y_0)-f_n(y_0)+2\epsilon\\
    &\leq \underset{y\in[a,b]}{\min}f_n(y)+3\epsilon.
\end{aligned}
\end{equation*}

Let $\epsilon\to 0$ and $n\to\infty$, we get
\begin{equation}\label{A3}
\underset{y\in[a,b]}{\min}f(y)\leq\lim\limits_{n\to\infty}\underset{y\in[a,b]}{\min}f_n(y).
\end{equation}

Combine (\ref{A2}) and (\ref{A3}), $\lim\limits_{n\to\infty}\underset{y\in[a,b]}{\min}f_n(y)=\underset{y\in[a,b]}{\min}f(y).$

\textbf{A2. Proof of Theorem \ref{thm01}\\}
\begin{romanlist}[(ii)]
\item For a given hyperparameter $\bm\theta\in\Theta$, $h\left(\bm x;\bm\omega,\bm\theta\right)\in\mathcal{Y}$ and $y\in\mathcal{Y}$ are both bounded, and $L$ satisfies the Lipschitz condition, so that
\begin{equation*}
\mathbb{E}\Big|L\Big(h(\bm x;\bm\omega,\bm\theta),y\Big)\Big|<\infty.
\end{equation*}

$(\bm x_{j,1}^{tr},y_{j,1}^{tr}),\cdots,(\bm x_{j,n_j^{tr}}^{tr},y_{j,n_j^{tr}}^{tr})$ are i.i.d.$\sim p_j(\bm x,y)$, Since $h$ and $L$ are continues and measurable, then for any given hyperparameter $\bm\theta\in\Theta$ and model parameter $\omega\in\Omega$,
$$L\Big(h(\bm x_{j,1}^{tr};\bm\omega,\bm\theta),y_{j,1}^{tr}\Big),\cdots,L\Big(h(\bm x_{j,n_j^{tr}}^{tr};\bm\omega,\bm\theta),y_{j,n_j^{tr}}^{tr}\Big)$$ are i.i.d.
According to the weak law of large numbers, when $n_j^{tr}\to\infty$,
\begin{equation*}
\frac{1}{n_j^{tr}}\sum_{i=1}^{n_j^{tr}}L\Big(h(\bm x_{j,i}^{tr};\bm\omega,\bm\theta),y_{j,i}^{tr}\Big)\overset{P}{\to}\mathbb{E}_{(\bm x,y)\sim p_j(\bm x,y)}L\Big(h(\bm x;\bm\omega,\bm\theta),y\Big).
\end{equation*}

For a given hyperparameter $\bm\theta\in\Theta$. Let
\begin{equation*}
\begin{aligned}
    f_{n_j^{tr}}(\bm\omega;D_j^{tr})&=\frac{1}{n_j^{tr}}\sum_{i=1}^{n_j^{tr}}L\Big(h(\bm x_{j,i}^{tr};\bm\omega,\bm\theta),y_{j,i}^{tr}\Big),\\
    f_j(\bm\omega)&=\mathbb{E}_{(\bm x,y)\sim p_j(\bm x,y)}L\Big(h(\bm x;\bm\omega,\bm\theta),y\Big).
\end{aligned}
\end{equation*}
Then $f_{n_j^{tr}}(\bm\omega;D_j^{tr})\overset{P}{\to}f_j(\bm\omega),\forall\bm\omega\in\Omega$. For any $\epsilon_1>0$ and $\delta_1>0$, let 
$$\tilde{G}_j=\cup\{D_j^{tr}:|f_{n_j^{tr}}(\bm\omega;D_j^{tr})-f_j(\bm\omega)|>\epsilon_1,D_j^{tr}\sim p_j(\bm x,y)\},$$
then $\exists N_1>0$, when $n_j^{tr}\geq N_1$, $\mu(\tilde{G}_j)<\delta_1$. So that when $D_j^{tr}\subset\tilde{G}^c_j$, $f_{n_j^{tr}}(\bm\omega;D_j^{tr})$ uniformly converges to $f_j(\bm\omega)$ on $\Omega$. By Lemma1 in appendix A1 we have
\begin{equation*}
\lim\limits_{n_j^{tr}\to\infty}\underset{\bm\omega\in\Omega}{\min}f_{n_j^{tr}}(\bm\omega;D_j^{tr})=\underset{\bm\omega\in\Omega}{\min}f_j(\bm\omega).
\end{equation*}

Since the minimum value of $L$ about $\bm\omega$ is unique, then for any hyperparameter $\bm\theta\in\Theta$, when $D_j^{tr}\subset\tilde{G}^c_j$ and $n_j^{tr}\to\infty$,
\begin{equation*}
\hat{\bm{\omega}}_j(\bm{\theta})=\underset{\bm{\omega}\in\Omega}{\arg\min}f_{n_j^{tr}}(\bm\omega)\to \underset{\bm{\omega}\in\Omega}{\arg\min}f_j(\bm\omega)={\bm{\omega}}_j(\bm{\theta}).
\end{equation*}

Since $h$ is continuous for $\bm\omega$ and $L$ is continuous, then when $D_j^{tr}\subset\tilde{G}^c_j$, for any $(\bm x_{j,i}^{val},y_{j,i}^{val})\in D_j^{val}$,
\begin{equation}\label{L_convergence}
\lim_{n_j^{tr}\to\infty}L\Big(h\Big(\bm{x}_{j,i}^{val};\hat{\bm{\omega}}_j(\bm{\theta}),\bm{\theta}\Big),y_{j,i}^{val}\Big)= L\Big(h\Big(\bm{x}_{j,i}^{val};\bm{\omega}_j(\bm{\theta}),\bm{\theta}\Big),y_{j,i}^{val}\Big).
\end{equation} 

According to Theorem 2 of~Ref.~\refcite{kanamori2012statistical}, we can obtain that when $n_j^{de},n_T\to\infty$, $\hat{r}_j(\cdot)\overset{P}{\to} r_j(\cdot)$, $j=1,2,\cdots,K.$ Here $\hat r_j(\cdot)$ is learned on $D_j^{de}$ and $D_T$. For any $\epsilon_2>0$ and $\delta_2>0$, let 
$$\hat{G}_j=\cup\{D_j^{de}:|\hat r_j(\bm x_{j,i}^{val})-r_j(\bm x_{j,i}^{val})|>\epsilon_2,D_j^{de}\sim p_j(\bm x,y)\},$$
then $\exists N_2>0$, when $\min(n_j^{de},n_T)\geq N_2$, $\mu(\hat{G}_j)<\delta_2$. Let $G_j=\tilde{G}_j^c\cap\hat{G}_j^c$, then $G^c_j=\tilde{G}_j\cup\hat{G}_j$ and 
\begin{equation}\label{Gj_measure}
\begin{aligned}
\mu(G^c_j)=1-\mu(G_j)<&1-\max(\mu(\tilde{G}_j^c),\mu(\hat{G}_j^c))\\
=&1-\max(1-\mu(\tilde{G}_j),1-\mu(\hat{G}_j))\\
=&\min(\mu(\tilde{G}_j),\mu(\hat{G}_j))\\
=&\min(\delta_1,\delta_2).
\end{aligned}
\end{equation}

For any given hyperparameter $\bm\theta\in\Theta$, from (\ref{L_convergence}) we obtain that for any $\epsilon_3>0$, $\exists N_3>0$, when $D_j^{tr}\subset G_j$ and $n_j^{tr}\geq N_3$, $$\Big|L\Big(h\Big(\bm{x}_{j,i}^{val};\hat{\bm{\omega}}_j(\bm{\theta}),\bm{\theta}\Big),y_{j,i}^{val}\Big)- L\Big(h\Big(\bm{x}_{j,i}^{val};\bm{\omega}_j(\bm{\theta}),\bm{\theta}\Big),y_{j,i}^{val}\Big)\Big|<\epsilon_3.$$

Then, for any $(\bm x,y)\in D_j^{val}$, when $D_j^{de}\subset G_j$ and $D_j^{tr}\subset G_j$, $\min(n_j^{de},n_T)>N_2$ and $n_j^{tr}>N_3$,
\begin{equation}\label{rL_convergence}
\begin{aligned}
    &\Big|\hat{r}_j(\bm x)\cdot L\Big(h\Big(\bm{x};\hat{\bm{\omega}}_j(\bm{\theta}),\bm{\theta}\Big),y\Big)-r_j(\bm x)\cdot L\Big(h\Big(\bm{x};\bm{\omega}_j(\bm{\theta}),\bm{\theta}\Big),y\Big) \Big|\\
    =&\Big|\hat{r}_j(\bm x)\cdot L\Big(h\Big(\bm{x};\hat{\bm{\omega}}_j(\bm{\theta}),\bm{\theta}\Big),y\Big)-r_j(\bm x)\cdot L\Big(h\Big(\bm{x};\hat{\bm{\omega}}_j(\bm{\theta}),\bm{\theta}\Big),y\Big)\\
    &+r_j(\bm x)\cdot L\Big(h\Big(\bm{x};\hat{\bm{\omega}}_j(\bm{\theta}),\bm{\theta}\Big),y\Big)
    -r_j(\bm x)\cdot L\Big(h\Big(\bm{x};\bm{\omega}_j(\bm{\theta}),\bm{\theta}\Big),y\Big) \Big|\\
    \leq&\Big|\hat{r}_j(\bm x)-r_j(\bm x)\Big|\cdot\Big| L\Big(h\Big(\bm{x};\hat{\bm{\omega}}_j(\bm{\theta}),\bm{\theta}\Big),y\Big)\Big|\\
    &+\Big|r_j(\bm x)\Big|\cdot\Big|L\Big(h\Big(\bm{x};\hat{\bm{\omega}}_j(\bm{\theta}),\bm{\theta}\Big),y\Big)-L\Big(h\Big(\bm{x};\bm{\omega}_j(\bm{\theta}),\bm{\theta}\Big),y\Big)\Big|\\
    \leq&\epsilon_2\Big| L\Big(h\Big(\bm{x};\hat{\bm{\omega}}_j(\bm{\theta}),\bm{\theta}\Big),y\Big)\Big|+C\cdot\epsilon_3.
\end{aligned}
\end{equation}

Let $\epsilon_2\to 0$ and $\epsilon_3\to 0$, then 
$$\Big|\hat{r}_j(\bm x)\cdot L\Big(h\Big(\bm{x};\hat{\bm{\omega}}_j(\bm{\theta}),\bm{\theta}\Big),y\Big)-r_j(\bm x)\cdot L\Big(h\Big(\bm{x};\bm{\omega}_j(\bm{\theta}),\bm{\theta}\Big),y\Big) \Big|\to 0.$$ 
Form (\ref{Gj_measure}), we can obtain that for any $\delta_4>0$ and $\epsilon_4>0$, $\exists N_4>0$, when $n_j^{de},n_j^{tr},n_T\geq N_4$, 
\begin{equation*}
\begin{aligned}
\mu\Big(\cup\Big\{D:&\Big|\hat{r}_j(\bm x)\cdot L\Big(h\Big(\bm{x};\hat{\bm{\omega}}_j(\bm{\theta}),\bm{\theta}\Big),y\Big)-r_j(\bm x)\cdot L\Big(h\Big(\bm{x};\bm{\omega}_j(\bm{\theta}),\bm{\theta}\Big),y\Big) \Big|\\
&>\epsilon_4,D_j^{de}\sim p_j(\bm x,y),D_j^{tr}\sim p_j(\bm x,y),D=D_j^{de}\cup D_j^{tr}\Big\}\Big)=\mu(G_j^c)<\delta_4.
\end{aligned}
\end{equation*}

Thus, for any given hyperparameter $\bm\theta\in\Theta$ and $(\bm x,y)\in D_j^{val}$, when $n_j^{de},n_j^{tr},n_T\to\infty$, 
\begin{equation}\label{rL}
\hat{r}_j(\bm x)\cdot L\Big(h\Big(\bm{x};\hat{\bm{\omega}}_j(\bm{\theta}),\bm{\theta}\Big),y\Big)\overset{P}{\to}r_j(\bm x)\cdot L\Big(h\Big(\bm{x};\bm{\omega}_j(\bm{\theta}),\bm{\theta}\Big),y\Big).
\end{equation}

So we have
\begin{equation}\label{A12}
\begin{aligned}
\lim\limits_{n_j^{de},n_j^{tr},n_T\to\infty}\hat{f}_{IW}(\bm\theta;D_j)=&\frac{1}{n_j^{val}}\sum_{i=1}^{n_j^{val}}\lim\limits_{n_j^{de},n_j^{tr},n_T\to\infty}\hat{r}_j(\bm x_{j,i}^{val})\cdot L\Big(h\Big(\bm{x}_{j,i}^{val};\hat{\bm{\omega}}_j(\bm{\theta}),\bm{\theta}\Big),y_{j,i}^{val}\Big)\\
=&\frac{1}{n_j^{val}}\sum_{i=1}^{n_j^{val}}r_j(\bm x_{j,i}^{val})\cdot L\Big(h\Big(\bm{x}_{j,i}^{val};\bm{\omega}_j(\bm{\theta}),\bm{\theta}\Big),y_{j,i}^{val}\Big),\\
\mathbb{E}\lim\limits_{n_j^{de},n_j^{tr},n_T\to\infty}\hat{f}_{IW}(\bm\theta;D_j)
=&\frac{1}{n_j^{val}}\sum_{i=1}^{n_j^{val}}\mathbb{E}r_j(\bm x_{j,i}^{val}) L\Big(h\Big(\bm{x}_{j,i}^{val};\bm{\omega}_j(\bm{\theta}),\bm{\theta}\Big),y_{j,i}^{val}\Big)\\
=&\mathbb{E}\frac{p_T(\bm x_{j,1}^{val})}{p_j(\bm x_{j,1}^{val})}L\Big(h\Big(\bm{x}_{j,1}^{val};\bm{\omega}_j(\bm{\theta}),\bm{\theta}\Big),y_{j,1}^{val}\Big)\\
=&\mathbb{E}\int_{\mathcal{X}}\frac{p_T(\bm x)}{p_j(\bm x)}L\Big(h\Big(\bm{x};\bm{\omega}_j(\bm{\theta}),\bm{\theta}\Big),y_{j,1}^{val}\Big)p_j(\bm x)d \bm x\\
=&\mathbb{E}\int_{\mathcal{X}}p_T(\bm x)L\Big(h\Big(\bm{x};\bm{\omega}_j(\bm{\theta}),\bm{\theta}\Big),y_{j,1}^{val}\Big)d\bm x\\
\overset{(*)}{=}&\mathbb{E}\int_{\mathcal{X}}L\Big(h\Big(\bm t;\bm{\omega}_j(\bm\theta),\bm\theta\Big),u\Big)p_T(\bm t)d\bm t\\
=&\mathbb{E}L\Big(h\Big(\bm t;\bm{\omega}(\bm{\theta}),\bm{\theta}\Big),u\Big)\\
=&f_T(\bm\theta).
\end{aligned}
\end{equation}
$(*)$ in (\ref{A12}) refers to~Ref.~\refcite{2000Improving} and \refcite{2007Covariate}. Since $\mathcal{Y}$ is bounded, $L$ satisfies the Lipschitz condition and $r(\cdot)\in[0,C]$ for a constant $C$, then $\lim\limits_{n_j^{de},n_j^{tr},n_T\to\infty}\hat{f}_{IW}(\bm\theta;D_j)$ is bounded.
Thus, by the bounded convergence theorem, we have
\begin{equation*}
\lim\limits_{n_j^{de},n_j^{tr},n_T\to\infty}\mathbb{E}\hat{f}_{IW}(\bm\theta;D_j)=\mathbb{E}\lim\limits_{n_j^{de},n_j^{tr},n_T\to\infty}\hat{f}_{IW}(\bm\theta;D_j)=f_T(\bm\theta).
\end{equation*}

\item The variance of $\hat f_{IW}$ is
\begin{equation*}
\begin{aligned}
\mathbb{V}\Big[\hat{f}_{IW}(\bm\theta;D_j)\Big]=&\mathbb{V}_{(\bm{x},y)\sim p_j(\bm x,y)}\Big[\hat{r}_j(\bm{x})L\Big(h\Big(\bm{x};\hat{\bm{\omega}}_j(\bm{\theta}),\bm{\theta}\Big),y\Big)\Big]\\
=&\mathbb{E}_{(\bm{x},y)\sim p_j(\bm x,y)}\Big[\hat{r}_j(\bm{x})L\Big(h\Big(\bm{x};\hat{\bm{\omega}}_j(\bm{\theta}),\bm{\theta}\Big),y\Big)\Big]^2\\
&-\Big\{\mathbb{E}_{(\bm{x},y)\sim p_j(\bm x,y)}\Big[\hat{r}_j(\bm{x})L\Big(h\Big(\bm{x};\hat{\bm{\omega}}_j(\bm{\theta}),\bm{\theta}\Big),y\Big)\Big]\Big\}^2.
\end{aligned}
\end{equation*}

By (\ref{rL}) and (\ref{A12}), according to the operation rules of limit and the bounded convergence theorem, we have
\begin{equation*}
\begin{aligned}
&\lim\limits_{n_j^{de},n_j^{tr},n_T\to\infty}\mathbb{E}_{(\bm{x},y)\sim p_j(\bm x,y)}\Big[\hat{r}_j(\bm{x})L\Big(h\Big(\bm{x};\hat{\bm{\omega}}_j(\bm{\theta}),\bm{\theta}\Big),y\Big)\Big]^2\\
=&\mathbb{E}_{(\bm{x},y)\sim p_j(\bm x,y)}\lim\limits_{n_j^{de},n_j^{tr},n_T\to\infty}\Big[\hat{r}_j(\bm{x})L\Big(h\Big(\bm{x};\hat{\bm{\omega}}_j(\bm{\theta}),\bm{\theta}\Big),y\Big)\Big]^2\\
=&\mathbb{E}_{(\bm{x},y)\sim p_j(\bm x,y)}\Big[\lim\limits_{n_j^{de},n_j^{tr},n_T\to\infty}\hat{r}_j(\bm{x})L\Big(h\Big(\bm{x};\hat{\bm{\omega}}_j(\bm{\theta}),\bm{\theta}\Big),y\Big)\Big]^2\\
=&\mathbb{E}_{(\bm{x},y)\sim p_j(\bm x,y)}\Big[r_j(\bm{x})L\Big(h\Big(\bm{x};\bm{\omega}_j(\bm{\theta}),\bm{\theta}\Big),y\Big)\Big]^2.
\end{aligned}
\end{equation*}
And by the same token, we can derive that
\begin{equation*}
\begin{aligned}
&\lim\limits_{n_j^{de},n_j^{tr},n_T\to\infty}\Big\{\mathbb{E}_{(\bm{x},y)\sim p_j(\bm x,y)}\Big[\hat{r}_j(\bm{x})L\Big(h\Big(\bm{x};\hat{\bm{\omega}}_j(\bm{\theta}),\bm{\theta}\Big),y\Big)\Big]\Big\}^2\\
=&\Big\{\lim\limits_{n_j^{de},n_j^{tr},n_T\to\infty}\mathbb{E}_{(\bm{x},y)\sim p_j(\bm x,y)}\Big[\hat{r}_j(\bm{x})L\Big(h\Big(\bm{x};\hat{\bm{\omega}}_j(\bm{\theta}),\bm{\theta}\Big),y\Big)\Big]\Big\}^2\\
=&\Big\{\mathbb{E}_{(\bm{x},y)\sim p_j(\bm x,y)}\lim\limits_{n_j^{de},n_j^{tr},n_T\to\infty}\Big[\hat{r}_j(\bm{x})L\Big(h\Big(\bm{x};\hat{\bm{\omega}}_j(\bm{\theta}),\bm{\theta}\Big),y\Big)\Big]\Big\}^2\\
=&\Big\{\mathbb{E}_{(\bm{x},y)\sim p_j(\bm x,y)}\Big[r_j(\bm{x})L\Big(h\Big(\bm{x};\bm{\omega}_j(\bm{\theta}),\bm{\theta}\Big),y\Big)\Big]\Big\}^2\\
=&f^2_T(\bm\theta).
\end{aligned}
\end{equation*}
Thus,
\begin{equation*}
\begin{aligned}
\lim\limits_{n_j^{de},n_j^{tr},\atop n_T\to\infty}\mathbb{V}\Big[\hat{f}_{IW}(\bm\theta;D_j)\Big]=&\mathbb{E}_{(\bm{x},y)\sim p_j(\bm x,y)}\Big[r_j(\bm{x})L\Big(h\Big(\bm{x};\bm{\omega}_j(\bm{\theta}),\bm{\theta}\Big),y\Big)\Big]^2-f^2_T(\bm\theta).
\end{aligned}
\end{equation*}
\end{romanlist}
\hfill\qed

\textbf{A2. Proof of Corollary \ref{cor01}\\}

\begin{romanlist}[(ii)]
\item
Since $\hat{r}_j(\cdot)\to r_j(\cdot)$ a.s. when $n_j^{de}\to\infty$ and $n_T\to\infty$, by (\ref{A11}) and the asymptotic unbiasedness of the sample variance and sample covariance, we can obtain that 
\begin{equation*}
\begin{aligned}
&\lim\limits_{n_j^{de},n_j^{tr},n_T\to\infty}\widehat{\operatorname{Var}}\Big[\hat{r}_{j}(\bm{x}_j^{val})\Big]=\operatorname{Var}_{\bm x\sim p_j(\bm x)}\Big[r_{j}(\bm{x})\Big],\\
&\lim\limits_{n_j^{de},n_j^{tr},n_T\to\infty}\widehat{\operatorname{Cov}}\Big[\hat{r}_{j}(\bm{x}_j^{val})\cdot L\Big(h\big(\bm x_j^{val};\hat{\bm{\omega}}_j(\bm{\theta}),\bm{\theta}\big),y_j^{val}\Big),\hat{r}_{j}(\bm{x}_j^{val})\Big]\\
=&\operatorname{Cov}_{(\bm x,y)\sim p_j(\bm x,y)}\Big[r_{j}(\bm{x})\cdot L\Big(h\big(\bm{x};\bm{\omega}_j(\bm{\theta}),\bm{\theta}\big),y\Big),r_{j}(\bm{x})\Big].
\end{aligned}
\end{equation*}

Because $Var_{\bm x\sim p_j(\bm x)}\left[r_{j}(\bm{x})\right]\neq 0$,
\begin{equation*}
\begin{aligned}
&\lim\limits_{n_j^{de},n_j^{tr},n_T\to\infty}\frac{\widehat{\operatorname{Cov}}\Big[\hat{r}_{j}(\bm{x}_j^{val})\cdot L\Big(h\big(\bm{x}_j^{val};\hat{\bm{\omega}}_j(\bm{\theta}),\bm{\theta}\big),y_j^{val}\Big),\hat{r}_{j}(\bm{x}_j^{val})\Big]}{\widehat{\operatorname{Var}}\Big[\hat{r}_{j}(\bm{x}_j^{val})\Big]}\\
=&\frac{\lim\limits_{n_j^{de},n_j^{tr},n_T\to\infty}\widehat{\operatorname{Cov}}\Big[\hat{r}_{j}(\bm{x}_j^{val})\cdot L\Big(h\big(\bm{x}_j^{val};\hat{\bm{\omega}}_j(\bm{\theta}),\bm{\theta}\big),y_j^{val}\Big),\hat{r}_{j}(\bm{x}_j^{val})\Big]}{\lim\limits_{n_j^{de},n_j^{tr},n_T\to\infty}\widehat{\operatorname{Var}}\Big[\hat{r}_{j}(\bm{x}_j^{val})\Big]}\\
=&\frac{\operatorname{Cov}_{(\bm x,y)\sim p_j(\bm x,y)}\Big[r_{j}(\bm{x})\cdot L\Big(h\big(\bm{x};\bm{\omega}_j(\bm{\theta}),\bm{\theta}\big),y\Big),r_{j}(\bm{x})\Big]}{Var_{\bm x\sim p_j(\bm x)}\Big[r_{j}(\bm{x})\Big]}.
\end{aligned}
\end{equation*}
Thus, we obtain that $\lim_{n_j^{de},n_j^{tr},n_T\to\infty}\hat{\eta}_j(\bm\theta)=\eta_j(\bm\theta).$

Then, we have
\begin{equation*}
\begin{aligned}
&\lim\limits_{n_j^{de},n_j^{tr},n_T\to\infty}\hat{f}_{CV}(\bm\theta;D_j)\\
=&\lim\limits_{n_j^{de},n_j^{tr},\atop n_T\to\infty}\frac{1}{n_j^{val}}\sum_{i=1}^{n_j^{val}}\Big[\hat{r}_j(\bm{x}_{j,i}^{val})L\Big(h\Big(\bm{x}_{j,i}^{val};\hat{\bm{\omega}}_j(\bm{\theta}),\bm{\theta}\Big),y_{j,i}^{val}\Big)+\hat{\eta}_j(\bm\theta)\cdot\Big(\hat r_j(x_{j,i}^{val})-1\Big)\Big]\\
=&\frac{1}{n_j^{val}}\sum_{i=1}^{n_j^{val}}\lim\limits_{n_j^{de},n_j^{tr},\atop n_T\to\infty}\Big[\hat{r}_j(\bm{x}_{j,i}^{val})L\Big(h\Big(\bm{x}_{j,i}^{val};\hat{\bm{\omega}}_j(\bm{\theta}),\bm{\theta}\Big),y_{j,i}^{val}\Big)+\hat{\eta}_j(\bm\theta)\cdot\Big(\hat r_j(x_{j,i}^{val})-1\Big)\Big]\\
=&\frac{1}{n_j^{val}}\sum_{i=1}^{n_j^{val}}\Big[r_j(\bm{x}_{j,i}^{val})L\Big(h\Big(\bm{x}_{j,i}^{val};\bm{\omega}_j(\bm{\theta}),\bm{\theta}\Big),y_{j,i}^{val}\Big)+\eta_j(\bm\theta)\cdot\Big(r_j(x_{j,i}^{val})-1\Big)\Big].
\end{aligned}
\end{equation*}

Since $$\mathbb{E}r_j(\bm x)=\int_{\mathcal{X}}r_j(\bm x)p_j(\bm x)d\bm x=\int_{\mathcal{X}}\frac{p_T(\bm x)}{p_j(\bm x)}p_j(\bm x)d\bm x=\int_{\mathcal{X}}p_T(\bm x)d\bm x=1,$$we can derive that
\begin{equation}\label{A21}
\begin{aligned}
&\mathbb{E}\lim\limits_{n_j^{de},n_j^{tr},n_T\to\infty}\hat{f}_{CV}(\bm\theta;D_j)\\
=&\frac{1}{n_j^{val}}\sum_{i=1}^{n_j^{val}}\mathbb{E}\Big[r_j(\bm{x}_{j,i}^{val})L\Big(h\Big(\bm{x}_{j,i}^{val};\bm{\omega}_j(\bm{\theta}),\bm{\theta}\Big),y_{j,i}^{val}\Big)+\eta_j(\bm\theta)\cdot\Big(r_j(x_{j,i}^{val})-1\Big)\Big]\\
=&\mathbb{E}\Big[r_j(\bm{x}_{j,1}^{val})L\Big(h\Big(\bm{x}_{j,1}^{val};\bm{\omega}_j(\bm{\theta}),\bm{\theta}\Big),y_{j,1}^{val}\Big)+\eta_j(\bm\theta)\cdot\Big(r_j(x_{j,1}^{val})-1\Big)\Big]\\
=&\mathbb{E}\Big[r_j(\bm{x}_{j,1}^{val})L\Big(h\Big(\bm{x}_{j,1}^{val};\bm{\omega}_j(\bm{\theta}),\bm{\theta}\Big),y_{j,1}^{val}\Big)\Big]+\mathbb{E}\Big[\eta_j(\bm\theta)\cdot\Big(r_j(x_{j,1}^{val})-1\Big)\Big]\\
\overset{(**)}{=}&f_T(\bm\theta)+\eta_j(\bm\theta)\cdot\mathbb{E}\left[r_j(x_{j,1}^{val})-1\right]\\
=&f_T(\bm\theta).
\end{aligned}
\end{equation}
$(**)$ in (\ref{A21}) is derived from (\ref{A12}). Similar to the proof of Theorem \ref{thm01} and by the bounded convergence theorem, we have
$$\lim\limits_{n_j^{de},n_j^{tr},n_T\to\infty}\mathbb{E}\hat{f}_{CV} (\bm\theta;D_j)=\mathbb{E}\lim\limits_{n_j^{de},n_j^{tr},n_T\to\infty}\hat{f}_{CV}(\bm\theta;D_j)=f_T(\bm\theta).$$

\item
According to the operation rules of limit and the bounded convergence theorem, we have
\begin{equation}\label{A22}
\begin{aligned}
&\lim\limits_{n_j^{de},n_j^{tr},n_T\to\infty}\mathbb{V}\Big[\hat{f}_{CV}(\bm\theta;D_j)\Big]\\
=&\lim\limits_{n_j^{de},n_j^{tr},n_T\to\infty}\mathbb{E}\Big[\hat{f}_{CV}(\bm\theta;D_j)\Big]^2-\lim\limits_{n_j^{de},n_j^{tr},n_T\to\infty}\Big\{\mathbb{E}\Big[\hat{f}_{CV}(\bm\theta;D_j)\Big]\Big\}^2\\
=&\mathbb{E}\lim\limits_{n_j^{de},n_j^{tr},n_T\to\infty}\Big[\hat{f}_{CV}(\bm\theta;D_j)\Big]^2-\Big\{\lim\limits_{n_j^{de},n_j^{tr},n_T\to\infty}\mathbb{E}\Big[\hat{f}_{CV}(\bm\theta;D_j)\Big]\Big\}^2\\
=&\mathbb{E}\Big[\lim\limits_{n_j^{de},n_j^{tr},n_T\to\infty}\hat{f}_{CV}(\bm\theta;D_j)\Big]^2-f_T^2(\bm\theta)\\
=&\mathbb{E}\Big\{\frac{1}{n_j^{val}}\sum_{i=1}^{n_j^{val}}\Big[r_j(\bm{x}_{j,i}^{val})L\Big(h\Big(\bm{x}_{j,i}^{val};\bm{\omega}_j(\bm{\theta}),\bm{\theta}\Big),y_{j,i}^{val}\Big)+\eta_j(\bm\theta)\cdot\Big(r_j(\bm x_{j,i}^{val})-1\Big)\Big]\Big\}^2\\
&-f_T^2(\bm\theta)\\
=&\frac{1}{n_j^{val}}\Big\{\mathbb{E}_{(\bm x,y)\sim p_j(\bm x,y)}\Big[r_j(\bm{x})L\Big(h\Big(\bm{x};\bm{\omega}_j(\bm{\theta}),\bm{\theta}\Big),y\Big)+\eta_j(\bm\theta)\cdot\Big(r_j(\bm x)-1\Big)\Big]^2\\
&\quad\quad\quad-f_T^2(\bm\theta)\Big\}
\end{aligned}
\end{equation}

To derive the last equation of (\ref{A22}), let $Z,Z_1,\cdots,Z_n$ be i.i.d. random variables, 
\begin{equation*}
\begin{aligned}
&\mathbb{E}\Big[\frac{1}{n}\sum_{i=1}^{n}Z_i\Big]^2-(\mathbb{E}Z)^2
=\frac{1}{n^2}\mathbb{E}\Big[\sum_{i=1}^{n}Z_i^2+2\sum_{i<j}Z_i Z_j\Big]-(\mathbb{E}Z)^2\\
=&\frac{1}{n}\mathbb{E}Z^2+\frac{2}{n^2}\frac{n(n-1)}{2}(\mathbb{E}Z)^2-(\mathbb{E}Z)^2=\frac{1}{n}[\mathbb{E}Z^2-(\mathbb{E}Z)^2].
\end{aligned}
\end{equation*}

By the same token, we have 
\begin{equation*}
\begin{aligned}
&\lim\limits_{n_j^{de},n_j^{tr},n_T\to\infty}\mathbb{V}\Big[\hat{f}_{IW}(\bm\theta;D_j)\Big]\\
=&\frac{1}{n_j^{val}}\Big\{\mathbb{E}_{(\bm x,y)\sim p_j(\bm x,y)}\Big[r_j(\bm{x})L\Big(h\Big(\bm{x};\bm{\omega}_j(\bm{\theta}),\bm{\theta}\Big),y\Big)\Big]^2-f^2_T(\bm\theta)\Big\}.
\end{aligned}
\end{equation*}

The control variate method\cite{Lemieux2017} implies that 
\begin{equation*}
\begin{aligned}
&\mathbb{E}\Big[r_j(\bm{x})L\Big(h\Big(\bm{x};\bm{\omega}_j(\bm{\theta}),\bm{\theta}\Big),y\Big)+\eta_j(\bm\theta)\cdot\Big(r_j(x)-1\Big)\Big]^2\\
\leq&\mathbb{E}\Big[r_j(\bm{x})L\Big(h\Big(\bm{x};\bm{\omega}_j(\bm{\theta}),\bm{\theta}\Big),y\Big)\Big]^2,
\end{aligned}
\end{equation*}
so we have $$\lim\limits_{n_j^{de},n_j^{tr},n_T\to\infty}\mathbb{V}\Big[\hat{f}_{CV}(\bm\theta;D_j)\Big]\leq\lim\limits_{n_j^{de},n_j^{tr},n_T\to\infty}\mathbb{V}\Big[\hat{f}_{IW}(\bm\theta;D_j)\Big].$$

\end{romanlist}
\hfill\qed

\textbf{A3. Proof of Corollary \ref{cor02}\\}

\begin{romanlist}[(ii)]
\item For a given hyperparameter $\bm\theta\in\Theta$ and any set of weights for sources $\bm\lambda=\{\lambda_1,\cdots,\lambda_K\}$,
\begin{equation*}
\lim\limits_{\forall j,n_j^{de},n_j^{tr},n_T\to\infty}\hat f_{\bm\lambda}\Big(\bm\theta;\{D_j\}_{j=1}^K\Big)=\sum_{j=1}^K\lambda_j n_j^{val}\cdot
\lim\limits_{n_j^{de},n_j^{tr},n_T\to\infty}\hat{f}_{CV}(\bm\theta;D_j).
\end{equation*}
Similar to the proof of Theorem \ref{thm01}, we can obtain that
\begin{equation*}
\begin{aligned}
&\lim\limits_{\forall j,n_j^{de},n_j^{tr},n_T\to\infty}\mathbb{E}\hat f_{\bm\lambda}\Big(\bm\theta;\{D_j\}_{j=1}^K\Big)
=\mathbb{E}\lim\limits_{\forall j,n_j^{de},n_j^{tr},n_T\to\infty}\hat f_{\bm\lambda}\Big(\bm\theta;\{D_j\}_{j=1}^K\Big)\\
=&\sum_{j=1}^K\lambda_j n_j^{val}\cdot
\mathbb{E}\lim\limits_{n_j^{de},n_j^{tr},n_T\to\infty}\hat{f}_{CV}(\bm\theta;D_j)
=\sum_{j=1}^K\lambda_j n_j^{val}\cdot f_T(\bm\theta)
=f_T(\bm\theta).
\end{aligned}
\end{equation*}

We can see that $\hat f_{FedDA}\left(\bm\theta;\{D_j\}_{j=1}^K\right)$ is a special case of $\hat f_{\bm\lambda}\left(\bm\theta;\{D_j\}_{j=1}^K\right)$ where $\lambda_j=\hat\lambda_j(\bm\theta),j=1,2,\cdots,K$, so
\begin{equation*}
\lim\limits_{\forall j,n_j^{de},n_j^{tr},n_T\to\infty}\mathbb{E}\hat f_{FedDA}\Big(\bm\theta;\{D_j\}_{j=1}^K\Big)=f_T(\bm\theta).
\end{equation*}

\item 
Because $K$ source datasets are independent and each source dataset is i.i.d., we have
\begin{equation*}
\mathbb{V}\Big[\hat f_{\bm\lambda}\Big(\bm{\theta};\{D_j\}_{j=1}^K\Big)\Big]
=\sum_{j=1}^K\mathbb{V}\Big[\lambda_j n_j^{val}\hat f_{CV}(\bm\theta;D_j)\Big]
=\sum_{j=1}^K\lambda_j^2(n_j^{val})^2\mathbb{V}\Big[\hat f_{CV}(\bm\theta;D_j)\Big].
\end{equation*}
According to the expression of $\operatorname{Div}_j\left(\bm\theta\right)$ and (\ref{A22}), we can obtain that $$\lim\limits_{n_j^{de},n_j^{tr},n_T\to\infty}\mathbb{V}\Big[\hat{f}_{CV}(\bm\theta;D_j)\Big]=\frac{1}{n_j^{val}}\operatorname{Div}_j\left(\bm\theta\right),$$
Then, by the operation rules of limit 
\begin{equation*}
\begin{aligned}
&\lim\limits_{\forall j,n_j^{de},n_j^{tr},\atop n_j^{val},n_T\to\infty}\mathbb{V}\Big[\hat f_{\bm\lambda}\Big(\bm{\theta};\{D_j\}_{j=1}^K\Big)\Big]\\
=&\lim\limits_{\forall j,n_j^{de},n_j^{tr},\atop n_j^{val},n_T\to\infty}\sum_{j=1}^K\lambda_j^2(n_j^{val})^2\lim\limits_{n_j^{de},n_j^{tr},n_T\to\infty}\mathbb{V}\Big[\hat f_{CV}(\bm\theta;D_j)\Big]\\
=&\lim\limits_{\forall j,n_j^{de},n_j^{tr},\atop n_j^{val},n_T\to\infty}\sum_{j=1}^K\lambda_j^2n_j^{val}\operatorname{Div}_j\left(\bm\theta\right).
\end{aligned}
\end{equation*}

To investigate the asymptotic variance of FedDAE, we first study the limit of $\hat\lambda_j(\bm\theta)$, $j=1,2,\cdots,K$. According to the operation rules of limit, we have
\begin{equation*}
\begin{aligned}
&\lim\limits_{n_j^{de},n_j^{tr},n_T\to\infty}\widehat{\operatorname{Div}}_j\left(\bm\theta\right)\\
=&\frac{1}{n_j^{val}}\sum_{i=1}^{n_j^{val}} \lim\limits_{n_j^{de},n_j^{tr},n_T\to\infty}\Big[\hat{r}_j(\bm{x}_{j,i}^{val}) \cdot L\Big(h\Big(\bm{x}_{j,i}^{val};\hat{\bm{\omega}}_j(\bm{\theta}),\bm{\theta}\Big), y_{j,i}^{val}\Big)\\
&\quad\quad\quad\quad\quad\quad\quad\quad\quad\quad+\hat{\eta}_j(\bm\theta)\cdot\Big(\hat{r}_j(\bm{x}_{j,i}^{val})-1\Big)\Big]^2\\
&-\lim\limits_{n_j^{de},n_j^{tr}, n_T\to\infty}\Big[\frac{1}{n_j^{val}}\sum_{i=1}^{n_j^{val}}\Big(\hat{r}_j(\bm{x}_{j,i}^{val}) \cdot L\Big(h\Big(\bm{x}_{j,i}^{val};\hat{\bm{\omega}}_j(\bm{\theta}),\bm{\theta}\Big), y_{j,i}^{val}\Big)\\
&\quad\quad\quad\quad\quad\quad\quad+\hat{\eta}_j(\bm\theta)\cdot\Big(\hat{r}_j(\bm{x}_{j,i}^{val})-1\Big)\Big)\Big]^2\\
=&\frac{1}{n_j^{val}}\sum_{i=1}^{n_j^{val}} \Big[r_j(\bm{x}_{j,i}^{val}) \cdot L\Big(h\Big(\bm{x}_{j,i}^{val};\bm{\omega}_j(\bm{\theta}),\bm{\theta}\Big), y_{j,i}^{val}\Big)+\eta_j(\bm\theta)\cdot\Big(r_j(\bm{x}_{j,i}^{val})-1\Big)\Big]^2\\
&-\Big[\frac{1}{n_j^{val}}\sum_{i=1}^{n_j^{val}}\Big(r_j(\bm{x}_{j,i}^{val}) \cdot L\Big(h\Big(\bm{x}_{j,i}^{val};\bm{\omega}_j(\bm{\theta}),\bm{\theta}\Big), y_{j,i}^{val}\Big)+\eta_j(\bm\theta)\cdot\Big(r_j(\bm{x}_{j,i}^{val})-1\Big)\Big)\Big]^2.
\end{aligned}
\end{equation*}
Since $\eta_j(\bm\theta)$ is bounded, we can use strong law of large numbers and the operation rules of limit to obtain that
\begin{equation*}
\begin{aligned}
&\lim\limits_{n_j^{de},n_j^{tr},n_j^{val}, n_T\to\infty}\widehat{\operatorname{Div}}_j\left(\bm\theta\right)\\
=&\lim\limits_{n_j^{val}\to\infty}\frac{1}{n_j^{val}}\sum_{i=1}^{n_j^{val}} \Big[r_j(\bm{x}_{j,i}^{val}) \cdot L\Big(h\Big(\bm{x}_{j,i}^{val};\bm{\omega}_j(\bm{\theta}),\bm{\theta}\Big), y_{j,i}^{val}\Big)\\
&\quad\quad\quad\quad\quad\quad\quad\quad+\eta_j(\bm\theta)\cdot\Big(r_j(\bm{x}_{j,i}^{val})-1\Big)\Big]^2\\
&-\lim\limits_{n_j^{val}\to\infty}\Big[\frac{1}{n_j^{val}}\sum_{i=1}^{n_j^{val}}\Big(r_j(\bm{x}_{j,i}^{val}) \cdot L\Big(h\Big(\bm{x}_{j,i}^{val};\bm{\omega}_j(\bm{\theta}),\bm{\theta}\Big), y_{j,i}^{val}\Big)\\
&\quad\quad\quad\quad\quad+\eta_j(\bm\theta)\cdot\Big(r_j(\bm{x}_{j,i}^{val})-1\Big)\Big)\Big]^2\\
=&\mathbb{E}_{(\bm x,y)\sim p_j(\bm x,y)}\Big[r_j(\bm{x}) \cdot L\Big(h\Big(\bm{x};\bm{\omega}_j(\bm{\theta}),\bm{\theta}\Big), y\Big)+\eta_j(\bm\theta)\cdot\Big(r_j(\bm{x})-1\Big)\Big]^2\\
&-\Big\{\lim\limits_{n_j^{val}\to\infty}\frac{1}{n_j^{val}}\sum_{i=1}^{n_j^{val}}\Big(r_j(\bm{x}_{j,i}^{val}) \cdot L\Big(h\Big(\bm{x}_{j,i}^{val};\bm{\omega}_j(\bm{\theta}),\bm{\theta}\Big), y_{j,i}^{val}\Big)\\
&\quad\quad+\eta_j(\bm\theta)\cdot\Big(r_j(\bm{x}_{j,i}^{val})-1\Big)\Big)\Big\}^2\\
=&\mathbb{E}_{(\bm x,y)\sim p_j(\bm x,y)}\Big[r_j(\bm{x}) \cdot L\Big(h\Big(\bm{x};\bm{\omega}_j(\bm{\theta}),\bm{\theta}\Big), y\Big)+\eta_j(\bm\theta)\cdot\Big(r_j(\bm{x})-1\Big)\Big]^2\\
&-\Big\{\mathbb{E}_{(\bm x,y)\sim p_j(\bm x,y)}\Big[r_j(\bm{x}) \cdot L\Big(h\Big(\bm{x};\bm{\omega}_j(\bm{\theta}),\bm{\theta}\Big), y\Big)+\eta_j(\bm\theta)\cdot\Big(r_j(\bm{x})-1\Big)\Big]\Big\}^2\\
=&\mathbb{E}_{(\bm x,y)\sim p_j(\bm x,y)}\Big[r_j(\bm{x}) \cdot L\Big(h\Big(\bm{x};\bm{\omega}_j(\bm{\theta}),\bm{\theta}\Big), y\Big)+\eta_j(\bm\theta)\cdot\Big(r_j(\bm{x})-1\Big)\Big]^2-f^2_T(\bm\theta)\\
=&\operatorname{Div}_j\left(\bm\theta\right).
\end{aligned}
\end{equation*}

According to the operation rules of limit, we can derive that
\begin{equation*}
\begin{aligned}
&\lim\limits_{\forall j,n_j^{de},n_j^{tr},n_j^{val},n_T\to\infty}\hat\lambda_j(\bm\theta)\\
=&\lim\limits_{\forall j,n_j^{de},n_j^{tr},n_j^{val},n_T\to\infty}\Bigg(\widehat{\operatorname{Div}}_j\left(\bm\theta\right) \sum_{j=1}^K \frac{n_j^{val}}{\widehat{\operatorname{Div}}_j\left(\bm\theta\right)}\Bigg)^{-1}\\
=&\Bigg(\lim\limits_{\forall j,n_j^{de},n_j^{tr},n_j^{val},n_T\to\infty}\widehat{\operatorname{Div}}_j\left(\bm\theta\right) \sum_{j=1}^K \frac{n_j^{val}}{\widehat{\operatorname{Div}}_j\left(\bm\theta\right)}\Bigg)^{-1}\\
=&\Bigg(\operatorname{Div}_j\left(\bm\theta\right) \sum_{j=1}^K \frac{n_j^{val}}{\operatorname{Div}_j\left(\bm\theta\right)}\Bigg)^{-1}\\
=&\lambda_j(\bm\theta).
\end{aligned}
\end{equation*}

And by the same token, we have
\begin{equation*}
\begin{aligned}
&\lim\limits_{\forall j,n_j^{de},n_j^{tr},n_j^{val},n_T\to\infty}\mathbb{V}\Big[\hat f_{FedDA}\Big(\bm{\theta};\{D_j\}_{j=1}^K\Big)\Big]\\
=&\lim\limits_{\forall j,n_j^{de},n_j^{tr},n_j^{val},n_T\to\infty}\sum_{j=1}^K\hat\lambda_j(\bm\theta)^2(n_j^{val})^2\mathbb{V}\Big[\hat f_{CV}(\bm\theta;D_j)\Big]\\
=&\lim\limits_{\forall j, n_j^{val}\to\infty}\sum_{j=1}^K\lambda_j^2(\bm\theta)n_j^{val}\operatorname{Div}_j\left(\bm\theta\right)\\
=&\lim\limits_{\forall j,n_j^{val}\to\infty}\sum_{j=1}^K\Bigg(\operatorname{Div}_j\left(\bm\theta\right) \sum_{j=1}^K \frac{n_j^{val}}{\operatorname{Div}_j\left(\bm\theta\right)}\Bigg)^{-2}n_j^{val}\operatorname{Div}_j\left(\bm\theta\right)\\
=&\lim\limits_{\forall j,n_j^{val}\to\infty}\Bigg(\sum_{j=1}^K \frac{n_j^{val}}{\operatorname{Div}_j\left(\bm\theta\right)}\Bigg)\Bigg(\sum_{j=1}^K \frac{n_j^{val}}{\operatorname{Div}_j\left(\bm\theta\right)}\Bigg)^{-2}\\
=&\lim\limits_{\forall j,n_j^{val}\to\infty}\Bigg(\sum_{j=1}^K \frac{n_j^{val}}{\operatorname{Div}_j\left(\bm\theta\right)}\Bigg)^{-1}.
\end{aligned}
\end{equation*}

Then, for any given set of weights of sources $\bm\lambda=\{\lambda_1,\cdots,\lambda_K\}$, we can use the Cauchy-Schwarz inequality to obtain that,
\begin{equation*}
\begin{aligned}
&1=\Bigg(\sum_{j=1}^K \lambda_j n_j^{val}\Bigg)^2\leq\Bigg(\sum_{j=1}^K\lambda_j^2n_j^{val}\operatorname{Div}_j\left(\bm\theta\right)\Bigg)\Bigg(\sum_{j=1}^K \frac{n_j^{val}}{\operatorname{Div}_j\left(\bm\theta\right)}\Bigg)\\
&\implies\Bigg(\sum_{j=1}^K \frac{n_j^{val}}{\operatorname{Div}_j\left(\bm\theta\right)}\Bigg)^{-1}\leq\Bigg(\sum_{j=1}^K\lambda_j^2n_j^{val}\operatorname{Div}_j\left(\bm\theta\right)\Bigg)\\
&\implies\lim\limits_{\forall j,n_j^{val}\to\infty}\Bigg(\sum_{j=1}^K \frac{n_j^{val}}{\operatorname{Div}_j\left(\bm\theta\right)}\Bigg)^{-1}\leq\lim\limits_{\forall j,n_j^{val}\to\infty}\Bigg(\sum_{j=1}^K\lambda_j^2n_j^{val}\operatorname{Div}_j\left(\bm\theta\right)\Bigg)\\
&\implies\lim\limits_{\forall j,n_j^{de},n_j^{tr},\atop n_j^{val},n_T\to\infty}\mathbb{V}\Big[\hat f_{FedDA}\Big(\bm{\theta};\{D_j\}_{j=1}^K\Big)\Big]\leq\lim\limits_{\forall j,n_j^{de},n_j^{tr},\atop n_j^{val},n_T\to\infty}\mathbb{V}\Big[\hat f_{\bm\lambda}\Big(\bm{\theta};\{D_j\}_{j=1}^K\Big)\Big].
\end{aligned}
\end{equation*}
\end{romanlist}
\hfill\qed

\textbf{A4. Proof of Corollary \ref{cor03}\\}

Let $\tilde{\bm\lambda}$ be a set of weights for sources with $\tilde{\lambda}_j=1/n^{val},j=1,2,\cdots,K$, it is easily to prove that $\tilde{\bm\lambda}$ satisfies $\tilde{\lambda}_j\geq 0,j=1,2,\cdots,K$ and $\sum_{j=1}^K \tilde{\lambda}_j n_j^{val}=1$. Then, by Corollary \ref{cor02} we can derive that
$$\lim\limits_{\forall j,n_j^{de},n_j^{tr},\atop n_j^{val},n_T\to\infty}\mathbb{V}\Big[\hat f_{FedDA}\Big(\bm{\theta};\{D_j\}_{j=1}^K\Big)\Big]\leq\lim\limits_{\forall j,n_j^{de},n_j^{tr},\atop n_j^{val},n_T\to\infty}\mathbb{V}\Big[\hat f_{\tilde{\bm\lambda}}\Big(\bm{\theta};\{D_j\}_{j=1}^K\Big)\Big],$$
where
$$\hat f_{\tilde{\bm\lambda}}\Big(\bm\theta;\{D_j\}_{j=1}^K\Big)=\sum_{j=1}^K \tilde{\lambda}_j n_j^{val}\hat f_{CV}(\bm\theta;D_j)=\frac{1}{n^{val}}\sum_{j=1}^K n_j^{val}\cdot\hat f_{CV}(\bm\theta;D_j).$$
According to (\ref{04}) and the inequality in Corollary \ref{cor01}, we have
$$\lim\limits_{\forall j,n_j^{de},n_j^{tr},\atop n_j^{val},n_T\to\infty}\mathbb{V}\Big[\hat f_{\tilde{\bm\lambda}}\Big(\bm{\theta};\{D_j\}_{j=1}^K\Big)\Big]\leq\lim\limits_{\forall j,n_j^{de},n_j^{tr},\atop n_j^{val},n_T\to\infty}\mathbb{V}\Big[\hat f_{FedIW}\Big(\bm{\theta};\{D_j\}_{j=1}^K\Big)\Big].$$
Thus, we obtain
$$\lim\limits_{\forall j,n_j^{de},n_j^{tr},\atop n_j^{val},n_T\to\infty}\mathbb{V}\Big[\hat f_{FedDA}\Big(\bm{\theta};\{D_j\}_{j=1}^K\Big)\Big]\leq\lim\limits_{\forall j,n_j^{de},n_j^{tr},\atop n_j^{val},n_T\to\infty}\mathbb{V}\Big[\hat f_{FedIW}\Big(\bm{\theta};\{D_j\}_{j=1}^K\Big)\Big].$$
\hfill\qed

\end{document}